\documentclass{article}

\usepackage[preprint]{neurips_2020}

\usepackage[utf8]{inputenc} 
\usepackage[T1]{fontenc}    
\usepackage{hyperref}       
\usepackage{url}            
\usepackage{booktabs}       
\usepackage{amsfonts}       
\usepackage{nicefrac}       
\usepackage{microtype}      
\usepackage{bbm}
\usepackage{tablefootnote}

\usepackage{amsmath,amsfonts,bm}
\usepackage{amsthm}

\def\eqref#1{equation~\ref{#1}}

\def\floor#1{\lfloor #1 \rfloor}
\def\1{\bm{1}}

\def\eps{{\epsilon}}

\DeclareMathAlphabet{\mathsfit}{\encodingdefault}{\sfdefault}{m}{sl}
\SetMathAlphabet{\mathsfit}{bold}{\encodingdefault}{\sfdefault}{bx}{n}

\newcommand{\E}{\mathbb{E}}

\newcommand{\R}{\mathbb{R}}

\DeclareMathOperator*{\argmax}{arg\,max}
\DeclareMathOperator*{\argmin}{arg\,min}

\newtheorem{theorem}{Theorem}
\newtheorem{lemma}{Lemma}

\usepackage{graphicx}
\usepackage{xcolor}
\usepackage{subcaption}
\usepackage[ruled,vlined]{algorithm2e}

\newcommand{\zk}[1]{}
\newcommand{\fw}[1]{}
\newcommand{\hfb}[1]{}
\newcommand{\nik}[1]{}

\DeclareMathOperator{\RED}{RED}

\title{Practical and sample efficient zero-shot HPO}

\author{%
  Fela Winkelmolen \\
  Amazon \\
  \texttt{<firstlastname>@gmail.com}\\
  \And
  Nikita Ivkin\\
  Amazon \\
  \texttt{ivkin@amazon.com}\\
  \And
  H. Furkan Bozkurt\\
  Amazon \\
  \texttt{bozkurh@amazon.com}\\
  \And
  Zohar Karnin\\
  Amazon \\
  \texttt{zkarnin@amazon.com}\\
}

\begin{document}

\maketitle

\begin{abstract}
Zero-shot hyperparameter optimization (HPO) is a simple yet effective use of 
  transfer learning for constructing a small list of   
  hyperparameter (HP) configurations that complement each other. That is to say, for any 
  given dataset, at least one of them is expected to perform well.  
Current techniques for obtaining this list are computationally expensive as they rely on running training jobs on a diverse collection of datasets and a large collection of randomly drawn HPs. This cost is especially problematic in environments where the space of HPs is regularly changing due to new algorithm versions, or changing architectures of deep 
networks.
We provide an overview of available approaches and introduce two novel techniques to handle the problem. The first is based on a surrogate model and adaptively chooses pairs of dataset, configuration to query. The second, for settings where finding, tuning and testing a surrogate model is problematic, is a multi-fidelity technique combining HyperBand with submodular optimization.
We benchmark our methods experimentally on five tasks (XGBoost, LightGBM, CatBoost, MLP and AutoML) and show significant improvement in accuracy compared to standard zero-shot HPO with the same training budget.
In addition to contributing new algorithms, we provide an extensive study of the zero-shot HPO technique resulting in (1) default hyper-parameters for popular algorithms that would benefit the community using them, (2) massive lookup tables to further the research of hyper-parameter tuning.
\end{abstract}

\section{Introduction}

Hyperparameter (HP) tuning can be described as a blackbox optimization problem, where the goal is to minimize an expensive to evaluate blackbox function $\ell: \Theta \rightarrow \R$, with $\ell$ mapping a HP configuration $\theta \in \Theta$ to its generalization loss $\ell(\theta)$, and $\Theta$ representing the configuration search space. Typically we do not have access to gradient information and desire evaluating $\ell$ a small number of times, as each evaluation corresponds to a costly training of a machine learning model.

Most state-of-the-art HPO tools such as SMAC~\citep{hutter2010sequential},
ParamILS~\citep{hutter2009paramils} and BOHB~\citep{falkner2018bohb} use
Bayesian Optimization (BO) as the core component of their algorithms, where the
blackbox function $\ell$ is modelled by a probabilistic surrogate model. After
each evaluation, the returned value is incorporated into the surrogate model, which is then used to determine the next HP configuration to evaluate. 


This approach has room for improvement in two aspects. First, if the surrogate
model is trained from scratch, this leads to poor behavior in the initial
stages, where not much is learned, and we are essentially querying random
configurations. This is a problem either when the HPO budget, meaning the number
of evaluations of $\ell$, is low, or when the HP space $\Theta$ is very rich, and it takes time to learn a reasonable surrogate for $\ell$. The second issue is that the BO approach is by definition sequential; this poses a problem when we wish to speed up the HPO process by using multiple machines. Although there are several adaptations of BO in the literature aimed to solve both of these issues, zero-shot HPO is a natural fit to address both. It requires learning a set of HP configurations offline based on a meta-collection of datasets, thus taking advantage of external information. Furthermore, by having a pre-determined set of configurations, the HPO procedure becomes embarrassingly parallel.

To obtain the set of configurations, previous papers~\citep{wistuba2015learning, wistuba2015sequential, pfisterer2018learning} run an offline meta-training job of the following nature.
Given a collection of $D$ datasets, choose $N$ configurations and evaluate them
on all $D$ datasets. Configurations are picked at random or according to some heuristic such as $eps$-net.
This offline process produces a performance table, $i,j$ entry of which
represents the loss of $i$-th configuration on $j$-th dataset. This performance table
is used to find a set of $K$ configurations that jointly minimize the aggregated loss over the whole dataset collection. 
In order to aggregate the losses over the datasets, it is common to  normalize the scores and minimize the average normalized score.

Notice that the offline procedure evaluates the function $\ell$ at $ND$
points, meaning it requires $ND$ training jobs. This might seem reasonable because it is a one-time procedure. 
However, we claim that that a more efficient procedure would be highly beneficial. 
First, algorithms have new versions with new hyper-parameters being introduced; 
the magnitude of the change in the $\ell$ function and training time is particularly prohibitive when deep networks are involved.
Second, even for a one-time job, the budget is limited, meaning that we have to
compromise on our choice of number of datasets $D$ or configurations $N$. In our experiments we verify that zero-shot HPO works better as we grow the number of datasets and the number of considered configurations.

To address this cost issue we design two solutions. The first is based on a surrogate function. The high level idea is that given the performance of several $i,j$ pairs, we can estimate the performance of new pairs. We design an adaptive method for selecting $i,j$ pairs to query while learning this surrogate function. This approach, called Offline Bayesian Optimization (OBO), adapts Bayesian optimization techniques to the non-standard optimization problem aimed at finding a set of HP configurations rather than an individual configuration. 

Although this method performs well, there is a non-trivial cost to it. Choosing the right surrogate model, and tuning it correctly involves a certain amount of manual labor, especially given that we have to do that as we query the $i,j$ pairs. In cases where this manual labor cannot be done, or a surrogate model is simply not good enough due to a particularly rich HP space, we provide a second solution. This second approach consists of a novel multi-fidelity algorithm solving the offline subset selection problem in a budget-efficient way. We evaluate both approaches in Section \ref{section:results}, demonstrating that their tradeoff of offline training budget $B$ vs.\ overall accuracy is superior to existing baselines.

All our experiments are conducted based on tables that we pre-computed for 5 machine learning problems, for a total of over 6 million training jobs. These tables and analysis done with them are worthy on their own. First, by publishing their content we hope to make HPO research more approachable to the scientific community. Second, the zero-shot configurations found by our experiments can be used to greatly improve the performance vs. run-time tradeoff for the algorithms considered in our study.

Another contribution, interesting on its own right, is that of the normalized score. Despite the abundance of papers evaluating a technique on multiple datasets, we find it surprising that there is no real standard for a normalized score allowing to aggregate an algorithm's performance across tasks. In Section~\ref{section:red} we formulate wanted properties of a normalized score. We review existing choices found in related literature and find that they do not have all of the wanted properties. For this reason we propose a normalized score called Relative Error Difference (RED), that obtains all needed properties. 

\section{Related work}

A number of previous works attempt to use evaluations of related tasks to speed up BO. \citet{feurer2015initializing} and \citet{brazdil2003ranking} propose to start the search from configurations that performed well on similar datasets and accomplish this by computing a distance metric on meta-data of the datasets. Although the results presented in the papers are positive, it is not always clear how to collect meta-data, which explicit meta-data features would be useful, and how to automatically get the correct distance metric relevant to the specific setting being solved.

\citet{wistuba2015learning} are the first to frame zero-shot HPO as an optimization problem minimizing the meta-loss over a collection of datasets. The HPs found this way are proposed as an initialization strategy for BO. Differentiable surrogate models are used to decrease the required evaluations on the meta-datasets. This, in combination with a discrete relaxation of the minimization problem, allows the authors to optimize the meta-loss using gradient descent. They show that using this initialization strategy followed by standard single task BO methods matches the performance of state-of-the-art meta-learning algorithms, as well as initialization strategies that make use of meta-data based distances. 

In a second paper~\citep{wistuba2015sequential}, the same authors use a greedy incremental algorithm to find a set of zero-shot HPO configurations, giving similar results. In this case, a full grid search is performed on all datasets, thus not requiring any surrogate model.
\citet{pfisterer2018learning} frame the zero-shot HPO problem as one of finding multiple default configurations. They combine both of the above ideas, by limiting the
search space to a discrete set of random configurations (as opposed to the grid search used in \citep{wistuba2015sequential}) and use an iterative greedy algorithm 
for finding a sequence of zero-shot configurations, but use a surrogate model for evaluations. They do not provide details about neither the type of surrogate model nor the amount of savings it brought. 

Similar ideas have been applied to other use cases. In particular, \citet{feurer2015initializing} use zero-shot HPO to select a portfolio of machine learning pipelines, while \citet{lindauer2018warmstarting} applies similar approaches to optimizing the configuration of SAT solvers. Finally, in \citet{van2018meta} each learned zero-shot hyperparameter is a function of the dataset meta-data, instead of being a fixed configuration.
\citet{perrone2019learning} make use of related tasks in order to limit the search space, rather than finding a small set of HP configurations. Specifically, they use the related tasks to eliminate regions of the search space that are very likely to provide poor results.

A different line of work attempts to use the evaluations of related tasks
directly in the modeling of the surrogate
function~\citep{yogatama2014efficient,perrone2018scalable}. However, developing
modeling techniques that robustly and effectively apply transfer learning on
real-world tasks has proven challenging, and thus is still an area under active research. 
See chapter 2 of \citet{hutter2019automl} for a broader review of such transfer
learning techniques. We note that (1) our technique is quite different and
likely complementary to that line of work, and (2) these works provide outside
information, but the problem of parallelism remains.
 

\section{Zero-shot HPO}

\begin{algorithm}[H]
  \SetAlgoLined
  \KwIn{number of desired zero-shot configurations $K$,\\ 
  ~~~~~~~~~~~~search space $\Theta$ containing $N$ configurations $\theta_1, ..., \theta_N$}
  \KwOut{$S \in [N]^K$, indices of found zero-shot configurations}
  
   \For{$j=1,2,\ldots,K$}{
     set $S_j = \arg \min_{i \in [N]} L(\{S_1,\ldots, S_{j-1}, i\})$
   }
   
   \caption{Naive greedy algorithm} \label{alg:naive}
  \end{algorithm}

Let us start by describing the computationally efficient greedy algorithm used by previous works. We will call this baseline Naive. We are given a collection of $D$ datasets, and a (possibly infinite) collection
$\Theta$ of configurations. In the naive approach, we select $N$  configurations $\theta_1,\ldots,\theta_N \in \Theta$ by i.i.d copies from a manually defined distribution. 
The amount of configurations $N$ is chosen according to the overall budget of training jobs. In this method we will invoke $ND$ training jobs, meaning that for a budget of $B$ we set $N=\floor{B/D}$.

We will use $[N]$ as a shorthand for the set of integers $\{1, 2, \ldots, N\}$. For each configuration $\theta$ and dataset $d$, we run a training job to
compute the validation loss $\ell(d,\theta)$. Given these values, we run an optimization procedure finding $K$ configurations $S \in [N]^K$ that jointly obtain the best performance on the dataset collection, by minimizing the following zero-shot meta-loss

\begin{equation} \label{eq:metaloss}
L(S) = D^{-1}\sum_{d=1}^D \min_{i \in S}\{ \ell(d,\theta_i) \}    
\end{equation}

Finding the optimal subset of size $K$ is known to be NP-hard. This was already pointed out in \citet{pfisterer2018learning}. Fortunately, there is a very simple approximation algorithm using the greedy approach. First, we show that $L$ is monotone decreasing
and supermodular. The proof of this lemma and the Theorem below is available in Appendix~\ref{ap:proofs}. 

\begin{lemma}\label{lemma:properties}
The function $L$ defined in Equation~\ref{eq:metaloss}
is monotone decreasing, i.e. $\forall A$ and $j\notin A$: $L(A\cup \{j\})\le L(A)$ 
and super-modular: $$\forall A \subseteq B \subset \mathbb{N} \text{ and } j \notin B:  
{L(A) - L(A\cup\{j\}) \ge L(B) - L(B\cup\{j\})}.$$
\end{lemma}

As a result of Lemma~\ref{lemma:properties} we know, by \cite{nemhauser1978analysis}, that the greedy solution aiming to minimize $L(S)$ provides a $(1-e^{-1})$-approximation to the optimal solution. For completeness, we provide an extension of this proof for the scenario where the local greedy step, i.e., the step finding the best configuration to add to the collection, is itself an approximation rather than an exact solution. This extension is useful for understanding the guarantees of methods that do not fill out the full $N \times D$ table.

\begin{theorem}\label{thm:aprox}
A greedy algorithm making steps towards $\eps$-approximates of best local improvements finds a $(1-e^{-1+\eps})$-approximation to the problem $\max_{|S|\le k} L(S)$ for monotone increasing submodular~$L$. 
\end{theorem}

\section{Offline Bayesian optimization (OBO)} \label{sec:obo}

When a probabilistic surrogate is available, an efficient zero-shot algorithm can be obtained by applying Bayesian optimization to optimize our zero-shot meta-loss. This is a fully sequential procedure to select evaluations, where at each time $t$ we select a candidate $\theta^t$ and a dataset $d^t$ to evaluate. 
We define $\ell^t(d,\theta)$ to be equal to $\infty$ if the configurations has not yet been selected at time $t$, and equal to $\ell(d,\theta)$ otherwise. 
In the same vein, $L^t(S) = D^{-1}\sum_{d=1}^D \min_{i \in S}\{ \ell^t(d,\theta_i) \}$, and at each time $t$ the current best zero-shot configurations are denoted with $S^t = \argmin_{S \in [N]^K} L^t(S)$.
At each step $t$ we train a probabilistic surrogate model $\mathcal{M}(\ell^t)$, that gives us a predictive distribution of $\ell$. We thus proceed by iteratively running the following two steps: (1) select the most promising configuration $\theta^t$, and its most promising location $l^t$, and (2) select the most promising dataset $d^t$ to evaluate using the configuration $\theta^t$ at location $l^t$.

Formally, at step $t$ we select the configuration $\theta^t$ and the location $l^t$ such that substituting location $l^t$ with $\theta^t$ and evaluating this configuration on all datasets brings the highest expected improvement, up to location $l^t$:

$$\theta^t, l^t = \argmax_{\theta, l \in \Theta\times[K]} \E_{\ell \sim \mathcal{M}(\ell^t)}\frac{
(\sum_{d=1}^D[\min_{i \in [l]}\ell^t(d,S^t_i)) - \min(\ell(d, \theta), \min_{i \in [l-1]} \ell^t(d,S^t_i))])^+
}{\sum_{d=1}^D\mathbbm{1}(\ell^t(d, \theta) = \infty)}$$

Note that any new candidate configuration will perform well only when enough datasets have been evaluated with this configuration. For this reason, we estimate the performance of a candidate configuration if we were to evaluate it on \emph{all} datasets. Otherwise we would never select new configurations. It is important to normalize by the number of remaining evaluations required to fully evaluate the configuration on all datasets. This way we select the configuration that maximizes the expected improvement \emph{per training job}.

Next, we select the dataset with the highest expected decrease in loss up to $l^t$, if $\theta^t$ is selected as the $l^t$-th zero-shot configuration:

$$d^t = \argmin_{d \in [D]} \E_{\ell \sim \mathcal{M}(\ell^t)}
[\min(\ell(d, \theta^t), \min_{i \in [l^t-1]} \ell^t(d,S^t_i)) - \min(\ell^t(d, \theta^t), \min_{i \in [l^t-1]}\ell^t(d,S^t_i)))]
$$

The above expectation is computed through sampling. As we choose to optimize for a specific location,  we do not need to compute the greedy procedure for each sample.

\paragraph{Choosing the surrogate model}
The first major decision regards the type of surrogate model. Here, although previous papers chose Gaussian Processes (GPs) we avoided this option since (1) our model must learn quickly (2) the hyper-parameter space can be quite rich, including categorical features that need special handling, and often conditional features that are known to interact poorly with GPs. For these reasons, in an attempt to obtain the best performing surrogate possible in our settings, we choose to use random forests as a surrogate model, as they have already been proven to work well on similar search spaces in \citet{hutter2010sequential}. 
The second major decision regards the input to the surrogate model. We experimented with (1) a single surrogate model taking in the dataset id as an input (2) separate models for each dataset (3) a mixture, where a global model takes in the predictions of the dataset specific model as inputs. Since a long portion of the time is spent where many datasets have very little information, we found that the variations including local models, meaning surrogate models for specific datasets, are outperformed by the first option having a single global model.


\section{Multi-fidelity (MF) zero-shot HPO} \label{sec:MF}

Let's begin with the setting of $K=1$. Here, the problem has a straightforward
reduction to the best arm identification task in multi-armed bandits. A query to
a configuration consists of selecting a random dataset and computing the loss
$\ell(d,\theta)$. This is an unbiased estimator of the mean over all datasets.
In our setting, we have a fixed budget of queries to $\ell$. As such, a first attempt would be to use the techniques provided by \citet{karnin2013almost} in their ``successive halving'' algorithm, proven to be highly efficient in the best arm identification task. Unfortunately, we have three added complications. First, we have a finite set of datasets, meaning that there is a maximum to the number of arm pulls. Second, when choosing an aggregation function other than mean, the confidence bounds become less trivial. This is mitigated by the analysis of \citet{jamieson2016non}, showing that the successive halving algorithm works well as the estimates converge to a single value, and there is no need for the algorithm to be aware of the convergence bound. Lastly, the number $N$ of possible configurations $\theta$ is not pre-determined as we can choose to explore either many configurations in a shallow manner or a few but exhaustively. This is exactly the problem dealt with by HyperBand~\citep{li2017hyperband}, where they build upon the successive halving algorithm.\looseness=-1

We conclude that for the setting of $K=1$ the problem fits the framework of
HyperBand (HB) exactly. Therefore we can use either HB or an extension of it, making use 
of surrogate functions such as BOHB~\citep{falkner2018bohb}. This solution for
$K=1$ suggests a solution for $K>1$
, where we aim to find a set $S$ of $K$ configurations.
We use the term \emph{location} for a chosen item in $S$ to 
denote its location in the sequential greedy process. With this term, the
algorithm chooses the item in location 1 based on the $\ell$ values, then chooses an item for 
location 2 based on its improvement over the $\ell$ value already achieved by location 1. It moves on to the other locations sequentially in the same way it handled 
location 2. We get an approximately optimal configuration for each location, thus by Theorem~\ref{thm:aprox} we get $(1 + e^{-1+\eps})$ approximation to the 
optimal subset $S$ of size $K$. 

Although this method is sensible at a high level there are a few key issues to be dealt with, that require a non-trivial solution.

{\bf Noise reduction}: When comparing two configurations based on a subset of the datasets of equal sizes we make sure the subsets contain the same datasets. This removes some variance from the estimates. 
To that end, we pick a single random shuffling of the datasets, and a query of configuration $\theta$ with resource $r$ means computing its loss on the first $r$ datasets and computing the mean. The same random order is used throughout the experiment

{\bf Information reuse}: Consider a setting where we wish to evaluate the
performance of a configuration $\theta$ as the second configuration in our set.
If $\theta$ happened to perform well as a first element, we queried its
performance on many datasets, and the evaluation actually comes for free. In general, when we start the process of discovering the configuration for location $>1$ in the set, we already have information from previous locations. We design the HyperBand instance to reuse this information. 

{\bf Resource Balancing}: It is unclear how we should balance the resources among the $K$ locations. Realistically, most of the value comes from the first locations. This can be made formal by noticing the magnitude of the losses and their variances become smaller as the location grows. 
As the magnitude of the loss shrinks, the benefit from minimizing it decreases, therefore the resources to locations with smaller losses can be smaller. 

The issue of resource balancing brings a major challenge. If we only start exploring location $j$ once we fixed locations $j'<j$, we cannot be adaptive. Moreover, even if we do not wish to be adaptive, by splitting the resources in advance, for example equally, we cannot control whether we end up actually using equal resources or, due to information reuse, end up with remaining resources and an unclear way to use them. The only straightforward way would be using them for the last location, but that is hardly a choice we would have done knowingly. To this end, we modify our HyperBand implementation in a way that allows exploring all locations simultaneously. 

The closest previous work we are aware of is that of \citet{streeter2009online}.
The authors provide an online algorithm that sequentially selects different sets
of size $K$ and, on average, competes with the optimal set in hindsight. Their technique provably works for any monotone decreasing supermodular function. This is a slightly different setting as the objective is to minimize regret rather than identifying the best subset. In order to handle the fact that a change in location $j$ changes the losses observed at locations $j'>j$, they require the algorithm for selecting each location to be very robust, and indeed they use EXP3.P~\citep{auer2002nonstochastic}, which is a multi-arm bandit algorithm for the regret setting, for adversarial realizations and a high probability guarantee. This algorithm is indeed robust enough to provide rigorous guarantees but not very practical for our setting. In particular, it is aimed for a regret guarantee rather than best arm identification. 

Due to this, we use an asynchronous HyberBand algorithm (see Appendix \ref{detailsmulti} for a more detailed description) with the following modification: at each step, when selecting what job to promote from a specific rung, we independently select the location that we are targeting. This location will determine the losses, and thus what job will be promoted to the next rung.
Thus, once the rung from which to promote is selected as usual, we select the new candidate to promote as analyzed in Algorithm~\ref{alg:hb selection}. We first select a location~$l$ such that the resources used per locations follows our predetermined allocation. Then, among all candidates in the current rung that have not yet been promoted, we select the candidate that gives the lowest loss when appended to the current best $l-1$ zero-shot candidates.

\begin{algorithm}[H]
\SetAlgoLined
\KwIn{Set $\Theta$ of candidates in the current rung, resource ratio vector $\rho$, previous resources used for each location $r$}
\KwOut{Next candidate $\theta_{\min}$}
	Let $l_{\max}$ be the smallest location index whose HB instance does not have a candidate evaluated on all datasets, or $1$ if no such index exists\;
	Let $l$ be $\argmin_{i \in [l_{\max}]} r_i \cdot \rho_i$\;
	Let $\theta_i, ..., \theta_{l-1}$ be the $l-1$ zero-zhot configurations as computed using Algorithm \ref{alg:naive} among all configurations already evaluated on all datasets\;
	Return $\argmin_{\theta \in \Theta} \sum_{d=1}^{r_l} \min(\min_{i \in [l-1]} \ell(d, \theta_i), \ell(d, \theta)$\;
 \caption{Anytime combinatorial HyperBand: candidate selection} \label{alg:hb selection}
\end{algorithm}



\section{Relative error difference (RED)}\label{section:red}

Zero-shot HPO can be used with any loss $\ell$ and any aggregate loss $L$. However, as we average losses computed over different datasets, one should be careful to choose a loss that can be meaningfully averaged across datasets.

The loss we use in our experiments is what we call \textit{relative error difference} (RED), computed comparing to a reference loss $r_{d}$ that we first compute for each dataset $d$. Thus, if $\tilde{\ell}(d, \theta)$ is an unnormalized error metric for dataset $d$, such as miss-classification rate, we normalize this metric by taking $\ell(d, \theta) = \RED(\tilde{\ell}(d, \theta), r_d)$, where

$$\RED(a, b) = \frac{a-b}{\max(a, b)}$$

here we assume $0/0$ to be equal to $0$.
As an example, if one dataset has a reference error rate of 40\%, and a second dataset has a reference error rate of 4\%, this metric considers a reduction of the former to 30\% equivalent to a reduction of the latter to 3\%.
RED provides us with robust aggregation thanks to its desirable properties compared to other normalization schemes, as described in more detail in Appendix \ref{ap:red}.


\section{Experimental setup}

\begin{table}[tbhp]
\centering
\caption{Size of pre-computed tables.}\label{tab:table sizes}
\begin{tabular}{lrrrr}
\toprule
setting & num. datasets & num. configs & num. HPs & total training jobs \\
\midrule
\textbf{XGBoost} & 80 & 18 406 & 9 & 1 472 480 \\
\textbf{LightGBM} & 88 & 30 000 & 9 & 2 640 000\\
\textbf{CatBoost} & 72 & 25 042 & 7 & 1 803 024\\
\textbf{MLP} & 59 & 5 445 & 8 & 321 255 \\
\textbf{ZAML} & 40 & 3 555 & 32 & 142 200 \\
\bottomrule 
\textbf{total}  & &  & & 6 378 959 \\
\end{tabular}
\end{table}

We test our approach on four commonly used general purpose supervised learning algorithms (XGBoost~\citep{chen2016xgboost}, CatBoost~\citep{prokhorenkova2018catboost}, LightGBM~\citep{ke2017lightgbm}, and MLP) and ZAML, a zero-shot AutoML search space, see Appendix~\ref{ap:searchspace} for details.
For each of the above algorithms we pre-compute the miss-classification rate for a large number of classification datasets using a large number of randomly sampled HP configurations.
We make the content of tables tables publicly available, with the hope it will aid further research into AutoML and transfer learning\footnote{Link will be made available in future versions}. For each training we store the validation and test miss-classification rate, as well as the wall-clock time required to complete the training job. We also provide the exact HPs used for each configuration.
All datasets used have at least $3000$ rows and originate from publicly available repositories: Kaggle\footnote{\url{https://kaggle.com}}, OpenML~\citep{OpenML2013}, UCI~\citep{Dua2019} and the AutoML
Challenge~\citep{automlchallenges}.
Minimal standard preprocessing, such as TF-IDF and one hot encoding, were applied to transform the input into purely numeric matrices.
To assure high generalization accuracy estimates, we use $.5/.25/.25$ sized splits for training, validation and test sets, respectively. 
The exact size of the tables and the number of HPs is listed in Table \ref{tab:table sizes}.
In the ZAML setting, we include the choice of feature processing and algorithm selection in the search space. Thus each configurations specifies the type of feature processing used, the HPs for the feature processors selected, the machine learning algorithm to use and the HPs of this algorithm. See Appendix \ref{ap:searchspace} for a full description of all search spaces used.

In all our experiments our loss is the average RED. The unnormalized error metric we use is the miss-classification rate, and our reference metric is 
obtained by averaging the test metric of the 10 models with the lowest validation error in our discrete set of random configurations. 
To use the limited number of datasets efficiently we evaluate our method in a leave-one-dataset-out fashion: when computing the performance of zero-shot configurations on any given datasets we use all other datasets as source tasks. We always report the \textit{test metric} of the hyperparameter configuration with the best \textit{validation error} up to any given number of evaluations.

We evaluate the following four algorithms, all with a budget of 3000 training jobs:

{\bf Naive}:  The baseline defined in Algorithm~\ref{alg:naive}.


{\bf Surrogate}: Both \citet{wistuba2015learning} and \citet{pfisterer2018learning} make use of surrogates trained on each dataset, to predict the performance of previously unseen configurations. The same considerations made in Section~\ref{sec:obo} for selecting the surrogate of OBO apply, thus we experimented with the same surrogate models. In addition, we attempted to use XGBoost instead of a random forest and use $1000$ trees in the joint models instead of the default $100$. The best performing surrogate, used for all further experiments, ended up being Scikit-learn's~\citep{scikit-learn} random forest with default HPs, with one surrogate trained per dataset.

{\bf OBO}: The BO approach of Section~\ref{sec:obo}. In our experiment this surrogate model is a random forest with $20$ trees. We consider the predictions from the single trees to be an approximation to taking $20$ samples from the posterior distribution. This is similar to what is done in SMAC~\citep{hutter2010sequential} for standard Bayesian optimization. As with {\bf Surrogate}, we used the default parameters provided by sklearn in order to avoid overfitting to our tables.

{\bf MF}: The multi-fidelity approach described in Section~\ref{sec:MF}.

We note that all preliminary comparisons for choosing the surrogate model were made on the XGBoost and MLP tables alone, while in all other settings we ran only the final version of the algorithms. This drastically reduces the chances that we suffered from overfitting to our experimental settings.

\section{Results}\label{section:results}


Our experiments show that the two algorithms proposed clearly outperform existing baselines. In table \ref{tab:obo} we can see that OBO consistently outperforms the Surrogate baseline. Similarly, as shown in Table \ref{tab:mf}, if a simpler method without surrogate is preferred, our multi-fidelity algorithm proves superior to the existing Naive approach used in previous works. Note that both tables display the RED between the given method and the baseline this method is compared to, as this is the most accurate way to compare two methods, yielding the most statistical power. In Appendix \ref{ap:results} we provide additional figures with comparisons to random search~\citet{bergstra2012random}, the online HPO technique that best competes with zero-shot HPO in terms of simplicity and parallelism. There we can see that random search requires a 10 to 50 times higher budget to match the performance of zero-shot HPO.
In the same appendix, we also provide additional results showing that in most settings OBO outperformed MF, however this gap disappeared for ZAML. Our conjecture is that when the search space becomes more complex and harder to model the benefit of a surrogate model disappears, an argument in favour of MF in these settings.


We hope that methods described, the zero-shot configurations published, and the tables we released will all help machine learning practitioners reach state-of-the-art model performance more quickly, freeing them from some of the costly and time consuming hyperparameter tuning effort normally required.

\begin{table}[tbhp]
\centering
\caption{OBO (vs Surrogate) -- avg RED ($\pm$std. err.) -- lower is better.}\label{tab:obo}
\begin{tabular}{lrrr}
\toprule
 & 1 zero-shot config                   & 2 zero-shot configs           & 5 zero-shot configs     \\
\midrule
\textbf{XGBoost} & -5.34\% ($\pm$ 1.17) & -5.07\% ($\pm$ 1.01) & -2.94\% ($\pm$ 0.92) \\
\textbf{LightGBM} & -0.74\% ($\pm$ 0.57) & -1.25\% ($\pm$ 0.58) & -0.94\% ($\pm$ 0.54) \\
\textbf{CatBoost} & -2.48\% ($\pm$ 1.55) & -1.69\% ($\pm$ 1.54) & -2.83\% ($\pm$ 1.58) \\
\textbf{MLP} & -9.09\% ($\pm$ 2.94) & -4.30\% ($\pm$ 1.54) & -1.60\% ($\pm$ 1.55) \\
\textbf{ZAML} & -1.24\% ($\pm$ 2.70) & -1.67\% ($\pm$ 3.07) & -0.62\% ($\pm$ 2.23) \\
\bottomrule 
\textbf{Combined} & -3.71\% ($\pm$ 0.77)  & -2.82\% ($\pm$ 0.63)  & -1.89\% ($\pm$ 0.57)  \\
\end{tabular}
\end{table}

\begin{table}[tbhp]
\centering
\caption{MF (vs Naive) -- avg RED ($\pm$std. err.) -- lower is better.}\label{tab:mf}
\begin{tabular}{lrrr}
\toprule
 & 1 zero-shot config                   & 2 zero-shot configs           & 5 zero-shot configs     \\
\midrule
\textbf{XGBoost} & -5.19\% ($\pm$ 1.75) & -5.66\% ($\pm$ 1.72) & -4.64\% ($\pm$ 1.71) \\
\textbf{LightGBM} & -4.68\% ($\pm$ 1.40) & -3.95\% ($\pm$ 1.45) & -4.16\% ($\pm$ 1.42) \\
\textbf{CatBoost} & +1.22\% ($\pm$ 0.87) & +0.56\% ($\pm$ 1.02) & +0.78\% ($\pm$ 0.99) \\
\textbf{MLP} & -1.22\% ($\pm$ 1.53) & -2.50\% ($\pm$ 1.17) & -1.81\% ($\pm$ 1.02) \\
\textbf{ZAML} & -0.43\% ($\pm$ 3.50) & -2.43\% ($\pm$ 3.70) & -0.95\% ($\pm$ 3.69) \\
\bottomrule 
\textbf{Combined} & -2.44\% ($\pm$ 0.77)  & -2.96\% ($\pm$ 0.77)  & -2.44\% ($\pm$ 0.76)  \\
\end{tabular}
\end{table}

\newpage
\section*{Broader Impact}

There has been a recent trend towards exponential growth in the compute resources required to obtain state-of-the-art results in machine learning\footnote{\url{https://openai.com/blog/ai-and-compute/}}, making research in some subfields prohibitively expensive. Though in general zero-shot HPO might benefit those with the resources to perform a expensive one-off offline optimization procedures, everyone can use the configurations computed with these techniques, assuming they are built for a publicly available algorithm. While on the one hand our work might increase the visibility and use of zero-shot HPO techniques, and this line of research might benefit those with access to massive computational resources, on the other hand this paper democratizes zero-shot HPO in three important ways: (1) reducing this offline cost, by providing two algorithm both explicitly aimed at reducing the cost of computing zero-shot HPO configurations (2), providing already computed zero-shot configurations for everyone to use on a selection of popular public algorithms, greatly reducing the difficulty and cost for practitioners to perform HPO, and (3) providing tables of pre-computed training results that can be used by researchers to benchmark new  HPO techniques without requiring prohibitive computational budgets.

\bibliographystyle{iclr2020_conference}
\bibliography{zeroshot_hpo.bib}

\begin{thebibliography}{29}
\providecommand{\natexlab}[1]{#1}
\providecommand{\url}[1]{\texttt{#1}}
\expandafter\ifx\csname urlstyle\endcsname\relax
  \providecommand{\doi}[1]{doi: #1}\else
  \providecommand{\doi}{doi: \begingroup \urlstyle{rm}\Url}\fi

\bibitem[Auer et~al.(2002)Auer, Cesa-Bianchi, Freund, and
  Schapire]{auer2002nonstochastic}
Peter Auer, Nicolo Cesa-Bianchi, Yoav Freund, and Robert~E Schapire.
\newblock The nonstochastic multiarmed bandit problem.
\newblock \emph{SIAM journal on computing}, 32\penalty0 (1):\penalty0 48--77,
  2002.

\bibitem[Bergstra \& Bengio(2012)Bergstra and Bengio]{bergstra2012random}
James Bergstra and Yoshua Bengio.
\newblock Random search for hyper-parameter optimization.
\newblock \emph{Journal of machine learning research}, 13\penalty0
  (Feb):\penalty0 281--305, 2012.

\bibitem[Brazdil et~al.(2003)Brazdil, Soares, and Da~Costa]{brazdil2003ranking}
Pavel~B Brazdil, Carlos Soares, and Joaquim~Pinto Da~Costa.
\newblock Ranking learning algorithms: Using ibl and meta-learning on accuracy
  and time results.
\newblock \emph{Machine Learning}, 50\penalty0 (3):\penalty0 251--277, 2003.

\bibitem[Chen \& Guestrin(2016)Chen and Guestrin]{chen2016xgboost}
Tianqi Chen and Carlos Guestrin.
\newblock Xgboost: A scalable tree boosting system.
\newblock In \emph{Proceedings of the 22nd acm sigkdd international conference
  on knowledge discovery and data mining}, pp.\  785--794, 2016.

\bibitem[Dua \& Graff(2017)Dua and Graff]{Dua2019}
Dheeru Dua and Casey Graff.
\newblock {UCI} machine learning repository, 2017.
\newblock URL \url{http://archive.ics.uci.edu/ml}.

\bibitem[Falkner et~al.(2018)Falkner, Klein, and Hutter]{falkner2018bohb}
Stefan Falkner, Aaron Klein, and Frank Hutter.
\newblock Bohb: Robust and efficient hyperparameter optimization at scale.
\newblock \emph{arXiv preprint arXiv:1807.01774}, 2018.

\bibitem[Feurer et~al.(2015)Feurer, Springenberg, and
  Hutter]{feurer2015initializing}
Matthias Feurer, Jost~Tobias Springenberg, and Frank Hutter.
\newblock Initializing bayesian hyperparameter optimization via meta-learning.
\newblock In \emph{Twenty-Ninth AAAI Conference on Artificial Intelligence},
  2015.

\bibitem[Guyon et~al.(2019)Guyon, Sun-Hosoya, Boull\'e, Escalante, Escalera,
  Liu, Jajetic, Ray, Saeed, Sebag, Statnikov, Tu, and Viegas]{automlchallenges}
Isabelle Guyon, Lisheng Sun-Hosoya, Marc Boull\'e, Hugo~Jair Escalante, Sergio
  Escalera, Zhengying Liu, Damir Jajetic, Bisakha Ray, Mehreen Saeed, Mich\'ele
  Sebag, Alexander Statnikov, WeiWei Tu, and Evelyne Viegas.
\newblock Analysis of the automl challenge series 2015-2018.
\newblock In \emph{AutoML}, Springer series on Challenges in Machine Learning,
  2019.
\newblock URL
  \url{https://www.automl.org/wp-content/uploads/2018/09/chapter10-challenge.pdf}.

\bibitem[Hutter et~al.(2019)Hutter, Kotthoff, and Vanschoren]{hutter2019automl}
F~Hutter, L~Kotthoff, and J~Vanschoren.
\newblock Automl: methods, systems, challenges (2018).
\newblock \emph{Book in preparation. Current draft at https://www. automl.
  org/book/. Accessed July}, 2019.

\bibitem[Hutter et~al.(2009)Hutter, Hoos, Leyton-Brown, and
  St{\"u}tzle]{hutter2009paramils}
Frank Hutter, Holger~H Hoos, Kevin Leyton-Brown, and Thomas St{\"u}tzle.
\newblock Paramils: an automatic algorithm configuration framework.
\newblock \emph{Journal of Artificial Intelligence Research}, 36:\penalty0
  267--306, 2009.

\bibitem[Hutter et~al.(2010)Hutter, Hoos, and
  Leyton-Brown]{hutter2010sequential}
Frank Hutter, Holger~H Hoos, and Kevin Leyton-Brown.
\newblock Sequential model-based optimization for general algorithm
  configuration (extended version).
\newblock \emph{Technical Report TR-2010--10, University of British Columbia,
  Computer Science, Tech. Rep.}, 2010.

\bibitem[Jamieson \& Talwalkar(2016)Jamieson and Talwalkar]{jamieson2016non}
Kevin Jamieson and Ameet Talwalkar.
\newblock Non-stochastic best arm identification and hyperparameter
  optimization.
\newblock In \emph{Artificial Intelligence and Statistics}, pp.\  240--248,
  2016.

\bibitem[Karnin et~al.(2013)Karnin, Koren, and Somekh]{karnin2013almost}
Zohar Karnin, Tomer Koren, and Oren Somekh.
\newblock Almost optimal exploration in multi-armed bandits.
\newblock In \emph{International Conference on Machine Learning}, pp.\
  1238--1246, 2013.

\bibitem[Ke et~al.(2017)Ke, Meng, Finley, Wang, Chen, Ma, Ye, and
  Liu]{ke2017lightgbm}
Guolin Ke, Qi~Meng, Thomas Finley, Taifeng Wang, Wei Chen, Weidong Ma, Qiwei
  Ye, and Tie-Yan Liu.
\newblock Lightgbm: A highly efficient gradient boosting decision tree.
\newblock In \emph{Advances in neural information processing systems}, pp.\
  3146--3154, 2017.

\bibitem[Li et~al.(2018)Li, Jamieson, Rostamizadeh, Gonina, Hardt, Recht, and
  Talwalkar]{li2018massively}
Liam Li, Kevin Jamieson, Afshin Rostamizadeh, Ekaterina Gonina, Moritz Hardt,
  Benjamin Recht, and Ameet Talwalkar.
\newblock Massively parallel hyperparameter tuning.
\newblock \emph{arXiv preprint arXiv:1810.05934}, 2018.

\bibitem[Li et~al.(2017)Li, Jamieson, DeSalvo, Rostamizadeh, and
  Talwalkar]{li2017hyperband}
Lisha Li, Kevin Jamieson, Giulia DeSalvo, Afshin Rostamizadeh, and Ameet
  Talwalkar.
\newblock Hyperband: A novel bandit-based approach to hyperparameter
  optimization.
\newblock \emph{The Journal of Machine Learning Research}, 18\penalty0
  (1):\penalty0 6765--6816, 2017.

\bibitem[Lindauer \& Hutter(2018)Lindauer and Hutter]{lindauer2018warmstarting}
Marius Lindauer and Frank Hutter.
\newblock Warmstarting of model-based algorithm configuration.
\newblock In \emph{Thirty-Second AAAI Conference on Artificial Intelligence},
  2018.

\bibitem[Nemhauser et~al.(1978)Nemhauser, Wolsey, and
  Fisher]{nemhauser1978analysis}
George~L Nemhauser, Laurence~A Wolsey, and Marshall~L Fisher.
\newblock An analysis of approximations for maximizing submodular set
  functions—i.
\newblock \emph{Mathematical programming}, 14\penalty0 (1):\penalty0 265--294,
  1978.

\bibitem[Pedregosa et~al.(2011)Pedregosa, Varoquaux, Gramfort, Michel, Thirion,
  Grisel, Blondel, Prettenhofer, Weiss, Dubourg, Vanderplas, Passos,
  Cournapeau, Brucher, Perrot, and Duchesnay]{scikit-learn}
F.~Pedregosa, G.~Varoquaux, A.~Gramfort, V.~Michel, B.~Thirion, O.~Grisel,
  M.~Blondel, P.~Prettenhofer, R.~Weiss, V.~Dubourg, J.~Vanderplas, A.~Passos,
  D.~Cournapeau, M.~Brucher, M.~Perrot, and E.~Duchesnay.
\newblock Scikit-learn: Machine learning in {P}ython.
\newblock \emph{Journal of Machine Learning Research}, 12:\penalty0 2825--2830,
  2011.

\bibitem[Perrone et~al.(2018)Perrone, Jenatton, Seeger, and
  Archambeau]{perrone2018scalable}
Valerio Perrone, Rodolphe Jenatton, Matthias~W Seeger, and C{\'e}dric
  Archambeau.
\newblock Scalable hyperparameter transfer learning.
\newblock In \emph{Advances in Neural Information Processing Systems}, pp.\
  6845--6855, 2018.

\bibitem[Perrone et~al.(2019)Perrone, Shen, Seeger, Archambeau, and
  Jenatton]{perrone2019learning}
Valerio Perrone, Huibin Shen, Matthias~W Seeger, Cedric Archambeau, and
  Rodolphe Jenatton.
\newblock Learning search spaces for bayesian optimization: Another view of
  hyperparameter transfer learning.
\newblock In \emph{Advances in Neural Information Processing Systems}, pp.\
  12751--12761, 2019.

\bibitem[Pfisterer et~al.(2018)Pfisterer, van Rijn, Probst, M{\"u}ller, and
  Bischl]{pfisterer2018learning}
Florian Pfisterer, Jan~N van Rijn, Philipp Probst, Andreas M{\"u}ller, and
  Bernd Bischl.
\newblock Learning multiple defaults for machine learning algorithms.
\newblock \emph{arXiv preprint arXiv:1811.09409}, 2018.

\bibitem[Prokhorenkova et~al.(2018)Prokhorenkova, Gusev, Vorobev, Dorogush, and
  Gulin]{prokhorenkova2018catboost}
Liudmila Prokhorenkova, Gleb Gusev, Aleksandr Vorobev, Anna~Veronika Dorogush,
  and Andrey Gulin.
\newblock Catboost: unbiased boosting with categorical features.
\newblock In \emph{Advances in neural information processing systems}, pp.\
  6638--6648, 2018.

\bibitem[Streeter \& Golovin(2009)Streeter and Golovin]{streeter2009online}
Matthew Streeter and Daniel Golovin.
\newblock An online algorithm for maximizing submodular functions.
\newblock In \emph{Advances in Neural Information Processing Systems}, pp.\
  1577--1584, 2009.

\bibitem[van Rijn et~al.(2018)van Rijn, Pfisterer, Thomas, Muller, Bischl, and
  Vanschoren]{van2018meta}
Jan~N van Rijn, Florian Pfisterer, Janek Thomas, Andreas Muller, Bernd Bischl,
  and Joaquin Vanschoren.
\newblock Meta learning for defaults--symbolic defaults.
\newblock In \emph{Neural Information Processing Workshop on Meta-Learning},
  2018.

\bibitem[Vanschoren et~al.(2013)Vanschoren, van Rijn, Bischl, and
  Torgo]{OpenML2013}
Joaquin Vanschoren, Jan~N. van Rijn, Bernd Bischl, and Luis Torgo.
\newblock Openml: Networked science in machine learning.
\newblock \emph{SIGKDD Explorations}, 15\penalty0 (2):\penalty0 49--60, 2013.
\newblock \doi{10.1145/2641190.2641198}.
\newblock URL \url{http://doi.acm.org/10.1145/2641190.2641198}.

\bibitem[Wistuba et~al.(2015{\natexlab{a}})Wistuba, Schilling, and
  Schmidt-Thieme]{wistuba2015learning}
Martin Wistuba, Nicolas Schilling, and Lars Schmidt-Thieme.
\newblock Learning hyperparameter optimization initializations.
\newblock In \emph{2015 IEEE international conference on data science and
  advanced analytics (DSAA)}, pp.\  1--10. IEEE, 2015{\natexlab{a}}.

\bibitem[Wistuba et~al.(2015{\natexlab{b}})Wistuba, Schilling, and
  Schmidt-Thieme]{wistuba2015sequential}
Martin Wistuba, Nicolas Schilling, and Lars Schmidt-Thieme.
\newblock Sequential model-free hyperparameter tuning.
\newblock In \emph{2015 IEEE international conference on data mining}, pp.\
  1033--1038. IEEE, 2015{\natexlab{b}}.

\bibitem[Yogatama \& Mann(2014)Yogatama and Mann]{yogatama2014efficient}
Dani Yogatama and Gideon Mann.
\newblock Efficient transfer learning method for automatic hyperparameter
  tuning.
\newblock In \emph{Artificial intelligence and statistics}, pp.\  1077--1085,
  2014.

\end{thebibliography}
\newpage

\appendix

\section{Comparison of RED to other metrics\label{ap:red}}
In this section we motivate our use of RED as an objective when computing the zero-shot configu\-rations, as well as when analyzing our results by aggregating the metric across multiple datasets. In particular, we motivate the use of RED over the following normalization schemes that have been used in the literature:

\begin{itemize}
  \item {\bf No normalization.} We denote the unnormalized miss-classification rate of the configuration~$\theta$ on dataset $d$ with $\tilde{\ell}(d, \theta)$.
  \item {\bf Rank normalization.} The rank normalized score is defined as the
      rank among $\Theta$, the set of all considered configurations, and can be normalized to be between 0 and 1: 
$$\ell(d, \theta) = \frac{|\{\tilde{\theta} \in \Theta~|~\tilde{\ell}(d, \tilde{\theta}) < \tilde{\ell}(d, \theta)\}|}{|\Theta|} $$
  \item {\bf Min-max normalization.} Linearly rescales the scores to keep the range between 0 and~1.
  \item {\bf Stddev normalization.} Linearly rescales the scores to have mean zero and standard de\-viation one. 
  \item {\bf RED.} The relative error difference metric described in section \ref{section:red}.
\end{itemize}

\citet{wistuba2015learning} use min-max normalization, whereas in \citet{wistuba2015sequential} min-max normalization is used only when evaluating the results, but rank normalization is used in the objective. And finally, \citet{pfisterer2018learning} use stddev normalization.

It is not immediately obvious what the trade-offs of different metrics are. We thus looked at the following two properties that we deem important when averaging metrics across datasets, and show that only RED has both:

\paragraph{Robust to rescaling.} When comparing metrics averaged over multiple datasets, without using any normalization, typically only the metrics with the largest range will meaningfully affect the results. While often the true utility of any improvement depends on the use case at hand, a desirable property---all other things being equal---is that a halving of the error metric always contributes the same value. Any of the normalization schemes described above, except for no normalization, are invariant under linear rescaling and thus have this property.
\paragraph{Robust to simple datasets.} There are some datasets for which in fact we do not want the results to affect the aggregate metric. These are datasets where the performance of the best configurations is very similar to the performance of the worst configuration. It is important to note that here we are talking about similarity in relative terms, not absolute terms. Reducing the error rate from 0.1\% to 0.001\% can be very valuable and will likely require a much better model. Conversely, reducing an error rate from 32\% to 31\% is, in most cases, not as impressive. All normalization schemes except RED and no normalization will magnify any difference in the case where all configurations perform very closely for a specific dataset. Thus only RED fulfills both desired properties.

\begin{table}[tbhp]
  \centering 
  \caption{Metric normalization comparison.}
  \begin{tabular}{lrrrrr}
  \toprule
  Normalization type &                none &           rank &          stddev &        min-max &           \color{red}{\textbf{RED}} \\
  \midrule
  Robust to rescaling &                no &             \textbf{yes} &  \textbf{yes} &  \textbf{yes} &      \textbf{yes} \\
  Robust to simple datasets &         \textbf{yes} &   no &            no &            no &      \textbf{yes} \\
  \bottomrule
  \end{tabular}
  \end{table}
  
Here we focused our discussion on the per dataset metric $\ell$, such that the values from different datasets can be aggregated. For the aggregation itself we use the average, as this gives us a principled way to approximate the expected metric on new datasets. There are however two cases where it might be desirable to explore alternative aggregation functions. The fist case is when we expect the new dataset to not come from (approximately) the same distribution as the meta datasets used to learn the zero-shot configurations. The second case is when we are not interested in the expected metric, for example if even in a single dataset the metric can vary by many orders of magnitude not proportional to the actual utility of the results. In such cases different aggregation functions such as median or P90 can be warranted. Our preliminary experiments (see Appendix \ref{ap:results}) however point towards the average to be a superior choice in our settings.

\section{Proofs}\label{ap:proofs}

For simplicity we redefine $D$ to be the set of all datasets.

\begin{proof}[\textbf{Proof of Lemma~\ref{lemma:properties}}]
    Let's first show that $L(S)$ is monotone decreasing. For any given $A$, $j$, and $d$:  
    \begin {align*} 
    L(A\cup \{j\}) &= \frac{1}{|D|}\sum_{d\in D} \min_{i \in A\cup \{j\}}\{
        \ell(d,\theta_i) \} \\
                    &= \frac{1}{|D|}\sum_{d\in D} \min\{\min_{i \in A}\{
\ell(d,\theta_i)\},  \ell(d,\theta_j)\} 
    \le \frac{1}{|D|}\sum_{d\in D} \min_{i \in A}\{\ell(d,\theta_i) \}  =   L(A)
    \end{align*}
    Next, we will show that $L(S)$ is super-modular. Recall that submodularity is implied by
    $$\forall A \subseteq B \subset \mathbb{N} \text{ and } j \notin B:  
{L(A) - L(A\cup\{j\}) \ge L(B) - L(B\cup\{j\})}$$
    Given any $A, B$ and $j$ we can split the
    dataset collection $D$ into three subsets such that $D = D_1 \cup D_2 \cup D_3$
\small    
$$D_1 = \left\{d \;| \ell(d, \theta_j) < \min_{i \in B}\{\ell(d,\theta_i) \}\right\},
\;\;  D_2 = \left\{d\; | \min_{i \in B}\{\ell(d,\theta_i) \} \le \ell(d,
\theta_j)  \le  \min_{i \in A}\{ \ell(d,\theta_i) \} \right\} $$
$$D_3 = \left\{d\; | \ell(d, \theta_j)  > \min_{i \in A}\{ \ell(d,\theta_i) \} \right\}$$
\normalsize    

Consider aggregate losses over these three collections defined as  
\small    
$$L_{D_i}(S) = \frac{1}{|D_i|}\sum_{d\in D_i} \min_{i \in S}\{ \ell(d,\theta_i)
\} \Rightarrow L_{D}(S) = \frac{1}{|D|} \left( |D_1|L_{D_1}(S) + |D_2|L_{D_2}(S) + |D_3|L_{D_3}(S)\right)$$
\normalsize    
Now, to show super-modularity of $L(S)$ it is sufficient to verity that for $i\in
\{1,2,3\}$ we have  
\begin{equation} \label{subcondition}
{L_{D_i}(A) - L_{D_i}(A\cup\{j\}) \ge L_{D_i}(B) - L_{D_i}(B\cup\{j\})}.
\end{equation}
By construction $L_{D_3}(A) = L_{D_3}(A\cup\{j\})$ and $L_{D_3}(B) =
L_{D_3}(B\cup\{j\})$, thus (\ref{subcondition}) holds for $i = 3$.
Further, $L_{D_2}(B) = L_{D_2}(B\cup\{j\})$ and $L_{D_2}(A) 
\ge L_{D_2}(A\cup\{j\})$ due to monotonicity, therefore (\ref{subcondition}) holds for $i = 2$.
For $i = 1$
\small
\begin{itemize}

\item $L_{D_1}(A) = L_{D_1}(A\cup\{j\}) + |D|^{-1}\sum_{d\in D_1} \left(\min_{i \in A}\{
\ell(d,\theta_i)\} - \ell(d,\theta_j)\right)$ 
\item $L_{D_1}(B) = L_{D_1}(B\cup\{j\}) + |D|^{-1}\sum_{d\in D_1} \left(\min_{i \in B}\{
\ell(d,\theta_i)\}- \ell(d,\theta_j)\right)$ 
\end{itemize}
\normalsize
Combining the two we get 
\small
$$L_{D_1}(A) - L_{D_1}(A\cup\{j\}) - L_{D_1}(B) + L_{D_1}(B\cup\{j\}) = 
|D|^{-1}\sum_{d\in D_1} \left(\min_{i \in A}\{\ell(d,\theta_i)\} - \min_{i \in B}\{
\ell(d,\theta_i)\}\right) \ge 0$$
\normalsize
where the last inequality holds due to $A \subseteq B \Rightarrow  \min_{i \in
    A}\{\ell(d,\theta_i)\} \ge \min_{i \in B}\{ \ell(d,\theta_i)\}$. 
    This confirms that (\ref{subcondition}) holds for $i=1$ and concludes the
    proof.

\end{proof}
\begin{proof}[\textbf{Proof of Theorem~\ref{thm:aprox}}]
Similarly to \cite{nemhauser1978analysis}, we prove by induction, however we slightly modify the invariant, which becomes
$$\forall i: L(opt) - L(S_i) \le \left(1- \frac{1-\eps}{k}\right)^i L(opt),$$
where $S_i$ is the set of items selected by the greedy algorithm after the $i$-th step, and $opt$ is the optimal set of $k$ items. It holds trivially for $i=0$, thus we assume it holds for step $i-1$ and prove the induction step. 
 Let the marginal improvement be defined as $\ell_{S_{i-1}}(j) = L(S_{i-1} \cup\{j\}) - L(S_{i-1})$, then, by submodularity, for each $j\in opt \setminus S_{i-1}$ we have
 \begin{align*}
    L(S_{i-1}\cup (opt\setminus S_{i-1})) &= L(S_{i-1}\cup (opt\setminus S_{i-1}\setminus \left\{j\right\})) +  \ell_{S_{i-1}\cup (opt\setminus S_{i-1}\setminus \left\{j\right\})}(j) \\ 
    &\le  L(S_{i-1}\cup (opt\setminus S_{i-1}\setminus \left\{j\right\})) +  \ell_{S_{i-1}}(j)
 \end{align*}
 Repeating the argument for all items $j\in opt\setminus S_{i-1}$ we obtain
 \begin{align*}
    L(S_{i-1}\cup (opt\setminus S_{i-1})) \le L(S_{i-1}) + \sum_{j\in opt\setminus S_{i-1}} \ell_{S_{i-1}}(j)
 \end{align*}
 And by monotonicity
 \begin{align} \label{eq:submod1}
    L(opt) \le
    L(S_{i-1}\cup (opt\setminus S_{i-1})) \le L(S_{i-1}) + \sum_{j\in opt\setminus S_{i-1}} \ell_{S_{i-1}}(j)
 \end{align}
 It is easy to show by contradiction that the item $j$ chosen at the $i$-th step has a marginal improvement
 \begin{align*}
    \ell_{S_{i-1}}(j) \ge \frac{1}{|opt\setminus S_{i-1}|} \sum_{\hat{j}\in opt\setminus S_{i-1}} \ell_{S_{i-1}}(\hat{j}) \ge \frac{1}{k}(L(opt) - L(S_{i-1}))
 \end{align*}
 Where the last inequality follows from Equation~\ref{eq:submod1}. However, our theorem states that the greedy algorithm will only find a $(1-\eps)$ approximation of the best marginal improvement, therefore  
 \begin{align*}
    \ell_{S_{i-1}}(j') \ge \frac{1 - \eps}{k}(L(opt) - L(S_{i-1}))
 \end{align*}
Which we can use to prove the induction step
\begin{align*}
 L(opt) - L(S_i) &= L(opt) - L(S_{i-1}) - \ell_{S_{i-1}}(j') \\
 &\le \left(L(opt) - L(S_{i-1})\right) - \frac{1-\eps}{k}\left(L(opt) - L(S_{i-1})\right)\\ 
 &\le \left(L(opt) - L(S_{i-1})\right) (1 - \frac{1-\eps}{k}) \le  (1 - \frac{1-\eps}{k})^i L(opt)
\end{align*}
 This completes the proof by induction. Applying the induction invariant to the $k$-th step we obtain our desired result
 $$L(opt) - L(S_k) \le \left(1 - \frac{1-\eps}{k}\right)^k L(opt) \le e^{-1+\eps} L(opt)$$
\end{proof}

\section{Details of anytime HyperBand\label{detailsmulti}}

Algorithm~\ref{alg:hb single} contains the pseudo-code for our anytime HB algorithm. Note that in our case the term resource used in the HB algorithm corresponds to the number of datasets. We always set the minimal resource to be 1.
Also, since we are choosing random datasets, this is a very crude but unbiased estimator of the mean. Querying a configuration with resource level $r$ means taking the average of training jobs of the HP configuration over $r$ datasets from our collection.
We split the possible resource into \emph{rungs}, in an exponential scale.
That is, for rung $s=0$, the resource level is $r=1$, for $s=1$, it is $r=\eta$ (default $3$), for 2 it is $r=\eta^2$, until $s=s_{\max}$ for which it is $r=D$, the number of datasets. To avoid technical complications and cumbersome notations we simply assume $D$ to be a power of $\eta$.
Algorithm~\ref{alg:hb single} describes the anytime HyperBand variant. Promoting a configuration to rung $s$ means running the HP configuration on $r=\eta^s$ datasets, or rather $\eta^s-\eta^{s-1}$ if we account for information reuse, in order to get a better estimate of its true value. The algorithm will initially query configurations in rung 0, and when it can, it will promote configurations to larger rungs. Unlike the ASHA algorithm in \cite{li2018massively}, that is an anytime parallel implementation of the simpler successive halving algorithm, Algorithm~\ref{alg:hb single} will once in a while choose not to promote a candidate from rung $s$ to rung $s+1$ but instead draw a new candidate and start it directly from rung $s+1$. This is exactly the idea of HyperBand that overcomes issues with regions in the configuration space that are `unlucky' in that they are misrepresented as high loss configurations when running with few resources. The exact ratio determining when a new configuration should be chosen rather than an old one to be promoted is set so that the overall resources used by each rung towards configurations that start in that rung (as opposed to being promoted to it) are equal. The expression of the ratio $\sum_{m=0}^{s-1} \frac{s_{\max} - s}{s_{max} - m}$ in Algorithm~\ref{alg:hb single} can be shown to provide an equal balance of resources. In other words, if we partition the configurations explored throughout the run of the algorithm according to the first rung they were launched (these are called brackets in HB), the sum of resources used by the configurations in a partition in the rung they started in should be the same. Since this is a trivial exercise we do not prove it.


\begin{algorithm}[H]
\SetAlgoLined
\KwIn{Configuration Selector $\Theta$, number of datasets $D$, elimination ratio $\eta$ (default value 3)}
\KwOut{Configuration $\theta$}
$s_{\max} = \log_\eta(D)$\;
\For{$t \in 1,2,\ldots$}{
	\For{$s=s_{\max}-1,s_{\max}-2,\ldots,0$}{
		\If{less than $1/\eta$ candidates in this rung $s$ has been promoted}{
			If we promoted over $\sum_{m=0}^{s-1} \frac{s_{\max} - s}{s_{max} - m}$ items from rung $s$, reset the counter of number of promoted items from rung $s$, select a new configuration $\theta_{\text{new}}$ from $\Theta$, and promote it to rung $s+1$ \;
			Otherwise, promote the best performing candidate $\theta$ to rung $s+1$ \;
		} 
	}
	If no item was promoted above, select a new configuration $\theta_{\text{new}}$ from $\Theta$ and run it in rung 0\;
	The best configuration $\theta$ at any time $t$ is the one with the minimal loss among those that ran with the full resources
 }
 
 \caption{Anytime HyperBand} \label{alg:hb single}
\end{algorithm}

\section{HP search space}\label{ap:searchspace}
For the hyper parameters search spaces for XGBoost, LightGBM, CatBoost and MLP see Table~\ref{table:searchspace}. 
Our zero-shot AutoML (ZAML) meta-pipeline has the following structure. First, a simple heuristic is applied to detect the type of each column: columns with less than 20 unique values are treated as categorical, columns containing entries from the list of 500 most frequent English words treated as text, and the remaining columns that can be converted to numeric value are labeled as numeric. All ZAML configurations use TF-IDF with fixed hyper parameters for preprocessing text features. For numeric features ZAML first fills in missing values, then applies one of four preprocessors: quantile transform, binning, standard scaler or min-max scaler. For categorical features ZAML applies either one-hot or ordinal encoding. Finally one of three algorithms is used to train on the preprocessed data: XGBoost, LightGBM or MLP. The HP search space for ZAML meta-pipeline is presented in Table~\ref{table:searchspace}.

\begin{table}[ht]
\centering 
\caption{HP distributions.}\label{table:searchspace}
\begin{tabular}{lrrr}
\toprule
Hyper parameter     &   Range                   &   Distribution\\
\midrule
\textbf{XGBoost} &~&~ \\  
n\_estimators       &   $[4, 512]$              &   log-uniform \\     
learning\_rate      &   $[10^{-6}, 1]$          &   log-uniform \\
gamma               &   $[2^{-20}, 64]$         &   log-uniform \\ 
min\_child\_weight  &   $[1, 32]$               &   log-uniform \\
max\_depth          &   $[2, 32]$               &   log-uniform \\
reg\_lambda         &   $[2^{-20}, 1]$          &   log-uniform \\
reg\_alpha          &   $[2^{-20}, 1]$          &   log-uniform \\
subsample           &   $[0.5, 1]$              &   uniform     \\
colsample\_bytree   &   $[0.3, 1]$              &   uniform     \\

\toprule
\textbf{LightGBM} &~&~ \\  
boosting\_type      &  ['gbdt', 'dart','goss']  & uniform\\
subsample           &  $[0.5, 1]$               & uniform\\
num\_leaves         &  $[30, 150]$              & uniform\\
learning\_rate      &  $[0.01, 0.1]$            & log-uniform\\
subsample\_for\_bin &  $[20000,30000]$          & uniform\\
min\_child\_samples &  $[20. 500]$              & uniform\\
reg\_alpha          &  $[0,1]$                  & uniform\\
reg\_lambda         &  $[0,1]$                  & uniform\\
colsample\_bytree   &  $[0.6,1]$                & uniform\\

\toprule
\textbf{CatBoost} &~&~ \\  
iterations             & $[10, 100]$ & uniform\\
depth                  & $[1, 8]$ & uniform\\
learning\_rate         & $[0.01, 1]$& log-uniform\\
random\_strength       & $[10^{-9}, 10]$ & log-uniform\\
bagging\_temperature   & $[0, 1]$ & uniform\\
border\_count          & $[1, 255]$ & uniform\\
l2\_leaf\_reg          & $[2, 30]$ & uniform\\

\toprule
\textbf{MLP} &~&~ \\  
batch\_size   & $[2^{4}, 10^{9}]$  & log\_uniform \\
lr           & $[10^{-4}, 10^{-1}]$  & log\_uniform \\
weight\_decay & $[10^{-5}, 10^{-1}]$ & log\_uniform \\
max\_units    & $[2^6, 2^{10}]$ & log\_uniform \\
momentum     & $[0.1, 0.9]$  & uniform \\
num\_layers   & $[1, 5]$      & uniform \\
max\_dropout  & $[0.1, 0.99]$ & uniform \\
use\_dropout  & [False, True] & uniform \\
activation   & ["relu"] & fixed \\
n\_epochs     & [100] & fixed \\
\toprule
\textbf{ZAML} &~&~ \\  
categorical preprocessor   & ["ordinal", "onehot"]  & uniform \\
numeric preprocessor       & ["quantile", "kbins", "standard", "minmax"]  & uniform \\
algorithm                  & ["MLP", "XGBoost", "LightGBM"] & uniform \\
kbins\_nbins               & [$2^4$, $2^8$]                     & log\_uniform \\
kbins\_encode              & ["onehot", "ordinal"]    &  uniform\\
quantile\_n\_quantiles     & [10, 1000]               &  log\_uniform\\
xgb\_hps                    & see above    & ~ \\
mlp\_hps                    & see above    & ~ \\
lgb\_hps                    & see above    & ~ \\

\bottomrule

\end{tabular}
\end{table}

\section{Zero-shot HP configurations\label{candidatestables}}

We report the 5 zero-shot configurations for each setting in Tables~\ref{tab:zero_shot_cand:xgb_amazon_v4}-\ref{tab:zero_shot_cand:fpp_amazon_v4}. These were found by applying the Naive algorithm on the full table of training results. These configurations can be used by users of these algorithms as a simple yet effective way to perform hyperparameter tuning and reproduce the results reported in this paper.

\begin{table}[!h]
\centering
\caption{Zero-shot configurations found for XGBoost.}
\label{tab:zero_shot_cand:xgb_amazon_v4}
\begin{tabular}{llllll}
\toprule
{} & zero-shot$_0$ & zero-shot$_1$ & zero-shot$_2$ & zero-shot$_3$ & zero-shot$_4$ \\
\midrule
colsample\_bytree  &      6.19e-01 &       8.4e-01 &      3.88e-01 &      8.79e-01 &      9.86e-01 \\
gamma              &      9.46e-05 &      7.04e-03 &      1.55e-05 &      8.43e-04 &      2.25e-05 \\
learning\_rate     &      6.74e-02 &      6.76e-02 &      9.74e-02 &      4.98e-03 &      3.57e-01 \\
max\_depth         &             7 &             3 &            18 &             9 &            20 \\
min\_child\_weight &             1 &             1 &             1 &             1 &             1 \\
n\_estimators      &           450 &           498 &           348 &           320 &           187 \\
reg\_alpha         &      3.69e-01 &      1.08e-04 &      6.57e-02 &      4.46e-03 &      2.97e-01 \\
reg\_lambda        &      6.08e-04 &      2.09e-05 &      1.05e-06 &      2.16e-04 &       3.8e-01 \\
subsample          &      8.29e-01 &       9.7e-01 &      5.25e-01 &      8.38e-01 &      9.04e-01 \\
\bottomrule
\\
\end{tabular}
\centering
\caption{Zero-shot configurations found for LightGBM.}
\label{tab:zero_shot_cand:lgb_amazon_v4}
\begin{tabular}{llllll}
\toprule
{} & zero-shot$_0$ & zero-shot$_1$ & zero-shot$_2$ & zero-shot$_3$ & zero-shot$_4$ \\
\midrule
boosting\_type      &          gbdt &          goss &          gbdt &          goss &          dart \\
colsample\_bytree   &      9.28e-01 &      9.38e-01 &      6.98e-01 &         8e-01 &      7.74e-01 \\
learning\_rate      &      9.47e-02 &       9.8e-02 &      8.86e-02 &      9.99e-02 &      2.71e-02 \\
min\_child\_samples &            27 &            57 &            21 &            34 &            21 \\
num\_leaves         &           149 &            35 &           101 &            92 &           120 \\
reg\_alpha          &       6.7e-02 &      3.99e-02 &      2.13e-01 &      4.49e-02 &      5.41e-01 \\
reg\_lambda         &      7.22e-01 &      1.13e-01 &       7.3e-01 &       2.3e-01 &      2.56e-01 \\
subsample           &       9.8e-01 &      8.65e-01 &      8.18e-01 &      6.34e-01 &       6.1e-01 \\
subsample\_for\_bin &         79750 &        106441 &        178491 &        272642 &        264651 \\
\bottomrule
\\
\end{tabular}
\centering
\caption{Zero-shot configurations found for CatBoost.}
\label{tab:zero_shot_cand:cat_amazon_v4}
\begin{tabular}{llllll}
\toprule
{} & zero-shot$_0$ & zero-shot$_1$ & zero-shot$_2$ & zero-shot$_3$ & zero-shot$_4$ \\
\midrule
bagging\_temperature &       9.4e-01 &      3.56e-01 &      9.21e-01 &      6.56e-01 &      7.12e-01 \\
border\_count        &           223 &            92 &           250 &            65 &            76 \\
depth                &             7 &             6 &             4 &             7 &             7 \\
iterations           &           714 &           992 &           837 &           870 &            92 \\
l2\_leaf\_reg        &             2 &             2 &            11 &             2 &             4 \\
learning\_rate       &       2.2e-01 &      1.75e-01 &      7.16e-02 &      1.88e-01 &      2.51e-02 \\
random\_strength     &      6.17e-05 &      4.08e-02 &      4.93e+00 &      1.72e+00 &      6.41e-08 \\
\bottomrule
\\
\end{tabular}
\centering
\caption{Zero-shot configurations found for MLP.}
\label{tab:zero_shot_cand:mlp_amazon_v4}
\begin{tabular}{llllll}
\toprule
{} & zero-shot$_0$ & zero-shot$_1$ & zero-shot$_2$ & zero-shot$_3$ & zero-shot$_4$ \\
\midrule
batch\_size   &           165 &            63 &            32 &            28 &            17 \\
lr            &      4.51e-02 &       1.3e-02 &      5.35e-02 &      7.77e-02 &      8.86e-02 \\
max\_dropout  &      9.35e-01 &      3.54e-01 &      4.87e-01 &      3.71e-01 &      2.53e-01 \\
max\_units    &           982 &            85 &           775 &           871 &           790 \\
momentum      &      8.86e-01 &      9.08e-01 &      8.75e-01 &      1.67e-01 &      8.18e-01 \\
num\_layers   &             3 &             2 &             4 &             4 &             4 \\
use\_dropout  &         False &         False &         False &         False &         False \\
weight\_decay &       4.3e-04 &      2.89e-03 &      3.21e-05 &       3.3e-05 &      1.19e-04 \\
\bottomrule
\end{tabular}
\end{table}

\begin{table}[!h]
\centering
\caption{Zero-shot configurations found for ZAML.}
\label{tab:zero_shot_cand:fpp_amazon_v4}
\begin{tabular}{llllll}
\toprule
{} & zero-shot$_0$ & zero-shot$_1$ & zero-shot$_2$ & zero-shot$_3$ & zero-shot$_4$ \\
\midrule
algo                      &           xgb &           mlp &           xgb &           mlp &           xgb \\
categorical               &        onehot &        onehot &        onehot &        onehot &        onehot \\
kbins\_encode             &        onehot &       ordinal &       ordinal &       ordinal &        onehot \\
kbins\_n\_bins            &            40 &            93 &            99 &            31 &            17 \\
numeric                   &      quantile &        minmax &        minmax &      standard &      quantile \\
quantile\_n\_quantiles    &           822 &           236 &           296 &            12 &            24 \\
mlp\_batch\_size          &            -- &            64 &            -- &            23 &            -- \\
mlp\_lr                   &            -- &      9.67e-02 &            -- &      4.53e-02 &            -- \\
mlp\_max\_dropout         &            -- &        8.e-01 &            -- &      8.85e-01 &            -- \\
mlp\_max\_units           &            -- &           460 &            -- &           978 &            -- \\
mlp\_momentum             &            -- &       7.6e-01 &            -- &      9.56e-01 &            -- \\
mlp\_num\_layers          &            -- &             3 &            -- &             4 &            -- \\
mlp\_use\_dropout         &            -- &         False &            -- &         False &            -- \\
mlp\_weight\_decay        &            -- &      1.81e-05 &            -- &        8.e-05 &            -- \\
xgb\_colsample\_bytree    &      5.52e-01 &            -- &      6.94e-01 &            -- &      8.11e-01 \\
xgb\_gamma                &      7.51e-04 &            -- &      8.82e-01 &            -- &      7.13e-06 \\
xgb\_learning\_rate       &       2.8e-01 &            -- &      3.73e-02 &            -- &      1.31e-06 \\
xgb\_max\_depth           &             5 &            -- &            16 &            -- &             6 \\
xgb\_min\_child\_weight   &             2 &            -- &             2 &            -- &             1 \\
xgb\_n\_estimators        &           195 &            -- &            94 &            -- &           252 \\
xgb\_reg\_alpha           &      5.23e-05 &            -- &      1.03e-03 &            -- &      1.11e-04 \\
xgb\_reg\_lambda          &      3.33e-06 &            -- &      2.35e-06 &            -- &      1.84e-06 \\
xgb\_subsample            &      8.76e-01 &            -- &      9.18e-01 &            -- &      8.97e-01 \\
\bottomrule
\end{tabular}
\end{table}

\section{Datasets\label{ap:datasets} and generated tables.}

All configurations were evaluated on the datasets listed in Table~\ref{table:list_datasets}. These are datasets taken from the following public repositories: Kaggle\footnote{\url{https://kaggle.com}}, OpenML~\citep{OpenML2013}, UCI~\citep{Dua2019} and the AutoML
Challenge~\citep{automlchallenges}.

As mentioned in the main text, we split every dataset in ratios 50/25/25 for training,
validation and test, respectively. Many datasets contain missing
values and non-numeric inputs, in these cases we apply basic feature type detection (numeric,
categorical and text) and preprocessing (one hot encoding for categorical features and
TF-IDF for text features). For each of the five settings considered we then start $N$ times $D$ training jobs, by evaluating $N$ randomly selected configurations on $D$ datasets. See Table \ref{tab:table sizes} for values of $N$ and $D$.

For each setting we make the following files available\footnote{Link will be made available in future versions}:

\textbf{error\_val.csv}: miss-classification rate on the validation set, stored in a matrix $M$ such that $M_{i,j}$ corresponds to the $j$-th dataset trained with the $i$-th configuration.

\textbf{error\_test.csv}: miss-classification rate on the test set, stored in a matrix $M$ such that $M_{i,j}$ corresponds to the $j$-th dataset trained with the $i$-th configuration.

\textbf{configurations.json}: a map from configuration index to the HPs of that configuration. Useful for training surrogate models.

\textbf{datasets.json}: mapping each dataset index to a dictionary containing the name and source of the dataset.

\begin{table}[tbhp]
\centering 
\caption{Datasets.}\label{table:list_datasets}
\begin{tabular}{lr}
\toprule
\textbf{UCI}:  
Abalone, Avila, BankMarketing, BlogFeedback, ChessKingRookvsKing, \\
ConditionBasedMaintenanceofNavalPropulsionPlants, \\
DatasetforSensorlessDriveDiagnosis, Diabetes130UShospitalsforyears19992008, \\
Dota2GamesResults, ElectricalGridStabilitySimulatedData, \\
FacebookCommentVolumeDataset, Gisette, HTRU2, IDA2016Challenge, \\
InsuranceCompanyBenchmarkCOIL2000, InternetAdvertisements, LetterRecognition, \\
MoCapHandPostures, Nursery, OnlineShoppersPurchasingIntentionDataset,\\
ParkinsonsTelemonitoring, PenBasedRecognitionofHandwrittenDigits, \\
PhysicalUnclonable, PhysicochemicalPropertiesofProteinTertiaryStructure, \\
PokerHand, SGEMMGPUkernelperformance, Spambase, StatlogLandsatSatellite,
\\
SuperconductivtyData, TurkiyeStudentEvaluation, UJIIndoorLoc, p53Mutants, \\
WeightLiftingExercisesmonitoredwithInertialMeasurementUnits, YearPredictionMSD \\

\midrule 
\textbf{AutoML Challenge}:
adult, albert, cadata, christine, digits, dilbert, dionis, fabert,\\ helena, jannis, madeline, philippine, robert, sylvine, volkert, yolanda \\
\midrule
\textbf{Kaggle}: 
blastchar/telco-customer-churn, burakhmmtgl/energy-molecule, \\
lpisallerl/air-tickets-between-shanghai-and-beijing, greenwing1985/housepricing,  \\
harlfoxem/housesalesprediction, jsphyg/weather-dataset-rattle-package, \\
lodetomasi1995/income-classification, loveall/appliances-energy-prediction, \\
contactprad/bike-share-daily-data, muonneutrino/us-census-demographic-data,\\
olgabelitskaya/classification-of-handwritten-letters, shrutimechlearn/churn-modelling, \\
umairnsr87/predict-the-number-of-upvotes-a-post-will-get \\
\midrule 
\textbf{OpenML}: 
wilt (40983), mfeat-morphological (18), ozone-level-8hr (1487), sick (38),\\ 
jm1 (1053), mfeat-fourier (14), har (1478), churn (40701), phoneme (1489), \\
wall-robot-navigation (1497),
numerai28.6 (23517), isolet (300), mfeat-pixel (40979), \\ first-order-theorem-proving (1475), 
mfeat-factors (12),
kc1 (1067), mnist\_784 (554), \\
GesturePhaseSegmentationProcessed (4538),
electricity (151), texture (40499),\\
jungle\_chess\_2pcs\_raw\_endgame\_complete (41027), PhishingWebsites (4534), \\ 
mfeat-karhunen (16), Bioresponse (4134),
connect-4 (40668), segment (40984), \\ mfeat-zernike (22), steel-plates-fault (40982), Fashion-MNIST (40996), nomao (1486), \\ splice (46), dna (40670), madelon (1485),\\
\bottomrule
\end{tabular}
\end{table}

\section{Additional results}\label{ap:results}

Tables \ref{tab:obo mf}-\ref{tab:obo naive} provide additional pairwise comparisons of the 4 zero-shot algorithms we tested. The RED in these tables is computed between the pair considered.

\begin{figure}[ht]
  \centering
  \begin{subfigure}[b]{0.40\linewidth}
    \includegraphics[width=\linewidth]{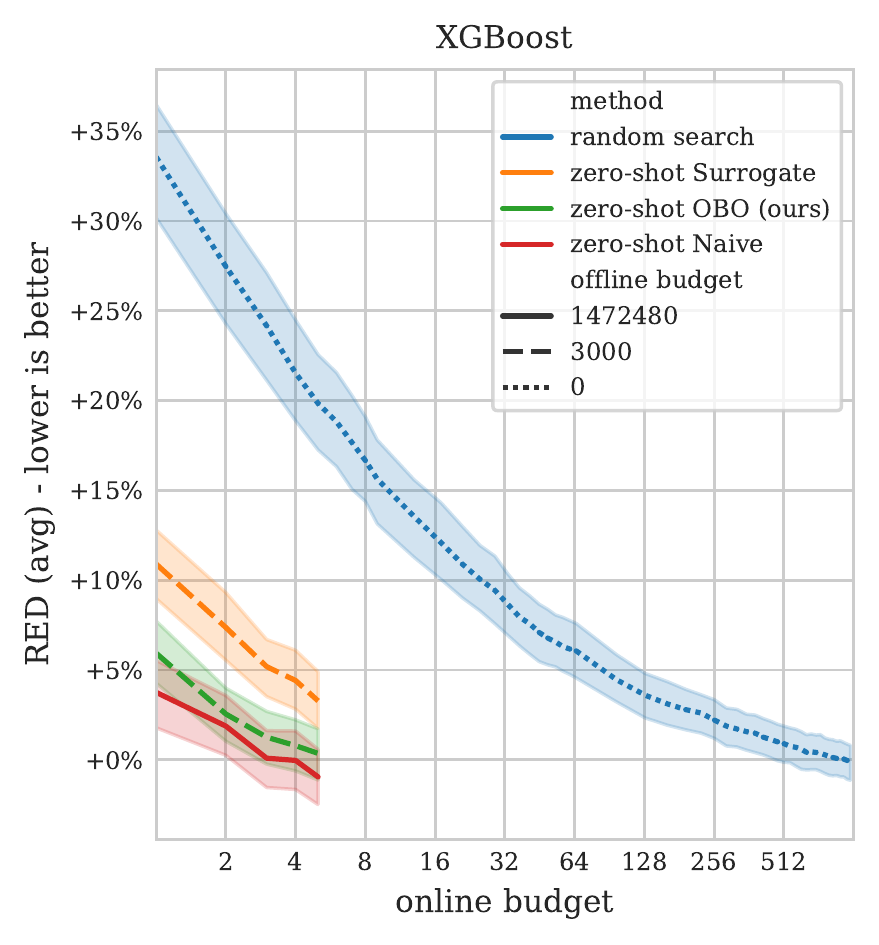}
  \end{subfigure}
  \begin{subfigure}[b]{0.40\linewidth}
    \includegraphics[width=\linewidth]{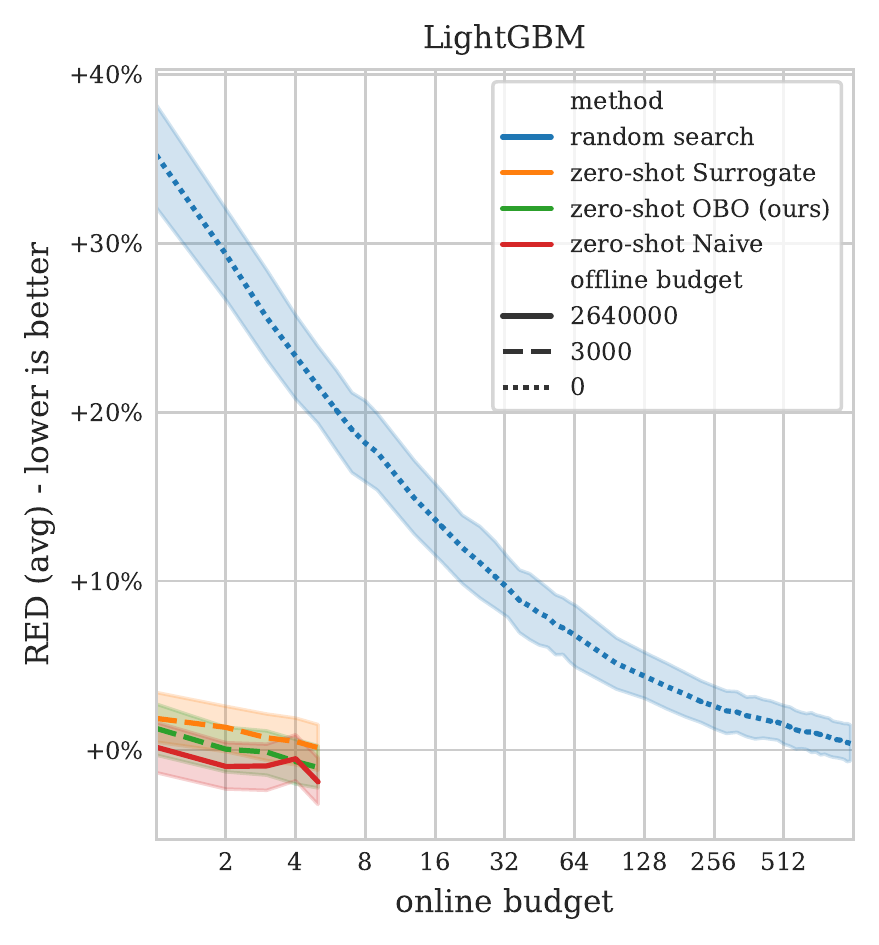}
  \end{subfigure}
  \begin{subfigure}[b]{0.40\linewidth}
    \includegraphics[width=\linewidth]{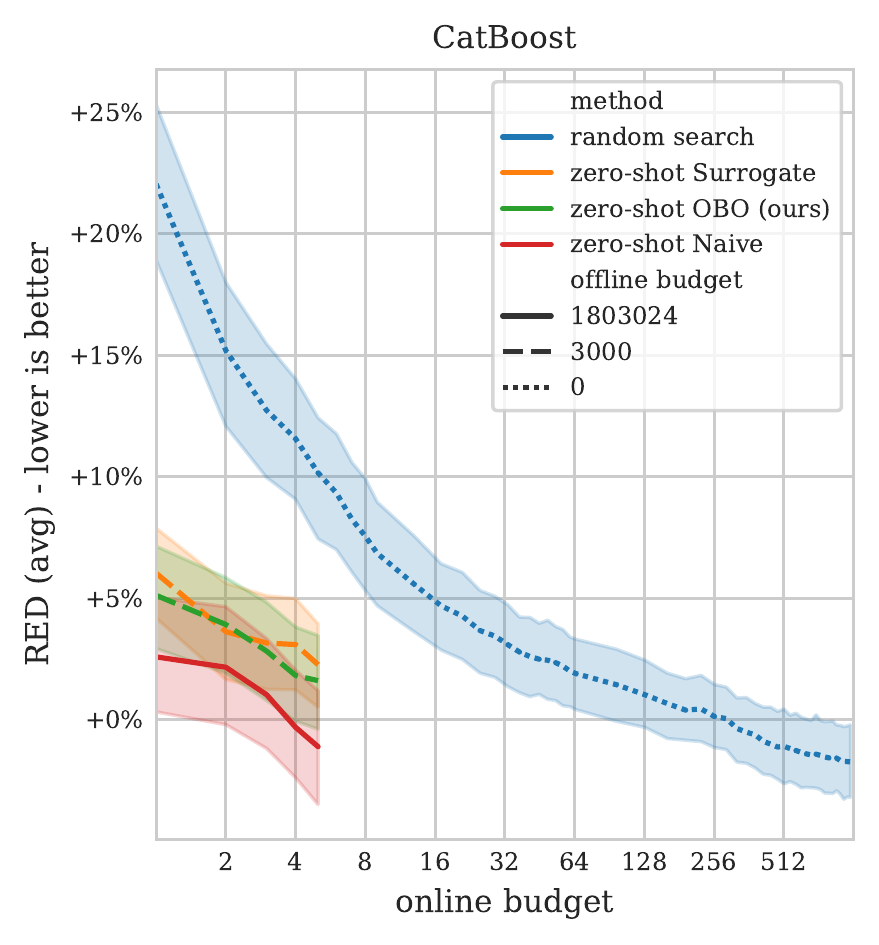}
  \end{subfigure}
  \begin{subfigure}[b]{0.40\linewidth}
    \includegraphics[width=\linewidth]{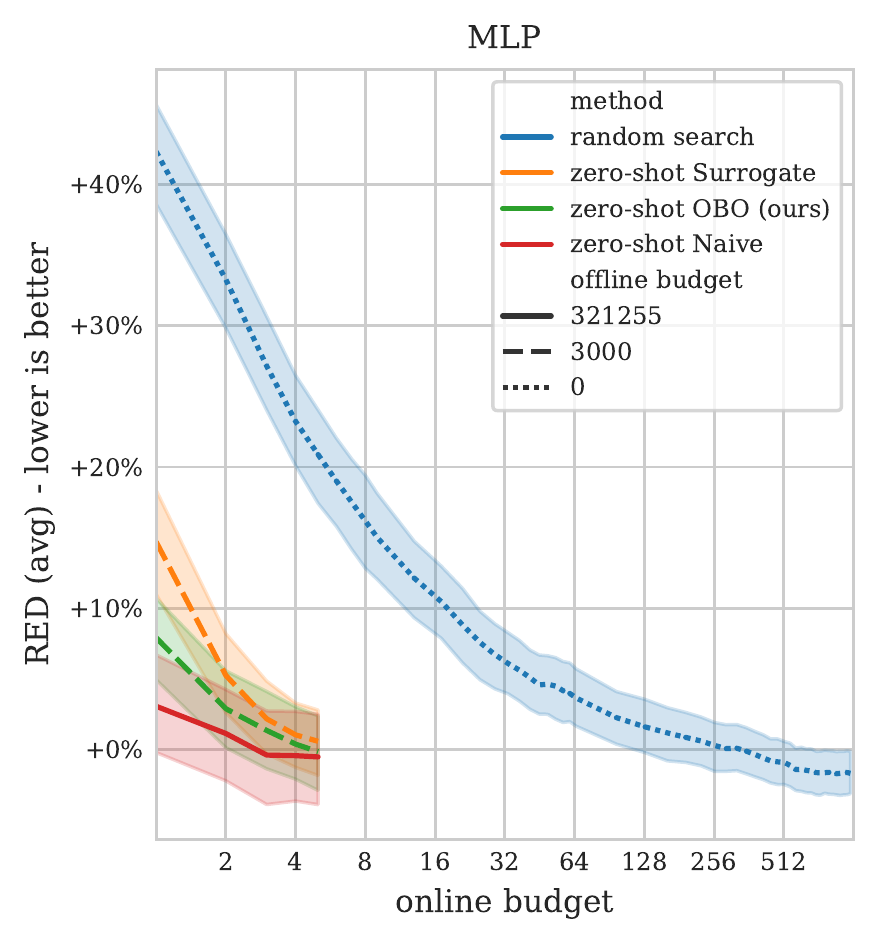}
  \end{subfigure}
  \begin{subfigure}[b]{0.40\linewidth}
    \includegraphics[width=\linewidth]{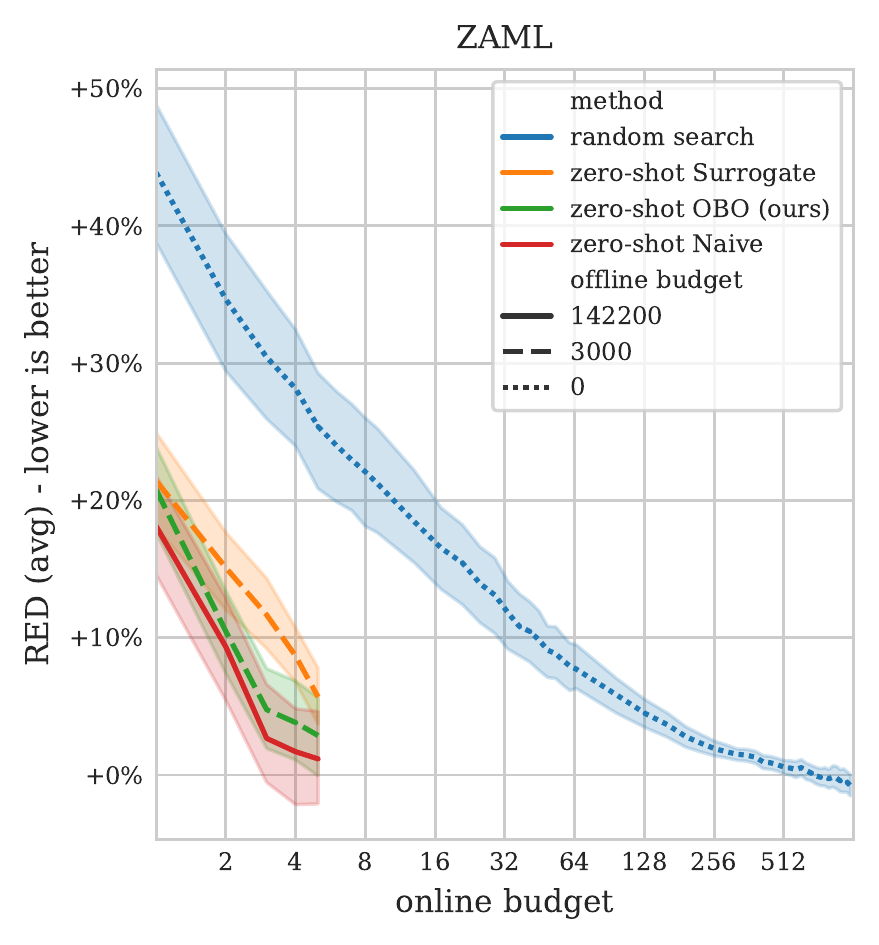}
  \end{subfigure}
  \caption{\textbf{Results for OBO}. Performance obtained by performing HPO using zero-shot configurations found by OBO (green). We compare the performance with the baseline (red) corresponding to zero-shot configurations computed applying the naive algorithm on all the data we have available (note the orders of magnitude more training jobs used to find these configurations), and Surrogate (orange), the best performing existing efficient method. For comparison we also include the performance of random search, which does not require any offline training, but as expected requires a much larger online budget to find competitive HP configurations.}
  \label{fig:obo}
\end{figure}

\begin{figure}[ht]
  \centering
  \begin{subfigure}[b]{0.40\linewidth}
    \includegraphics[width=\linewidth]{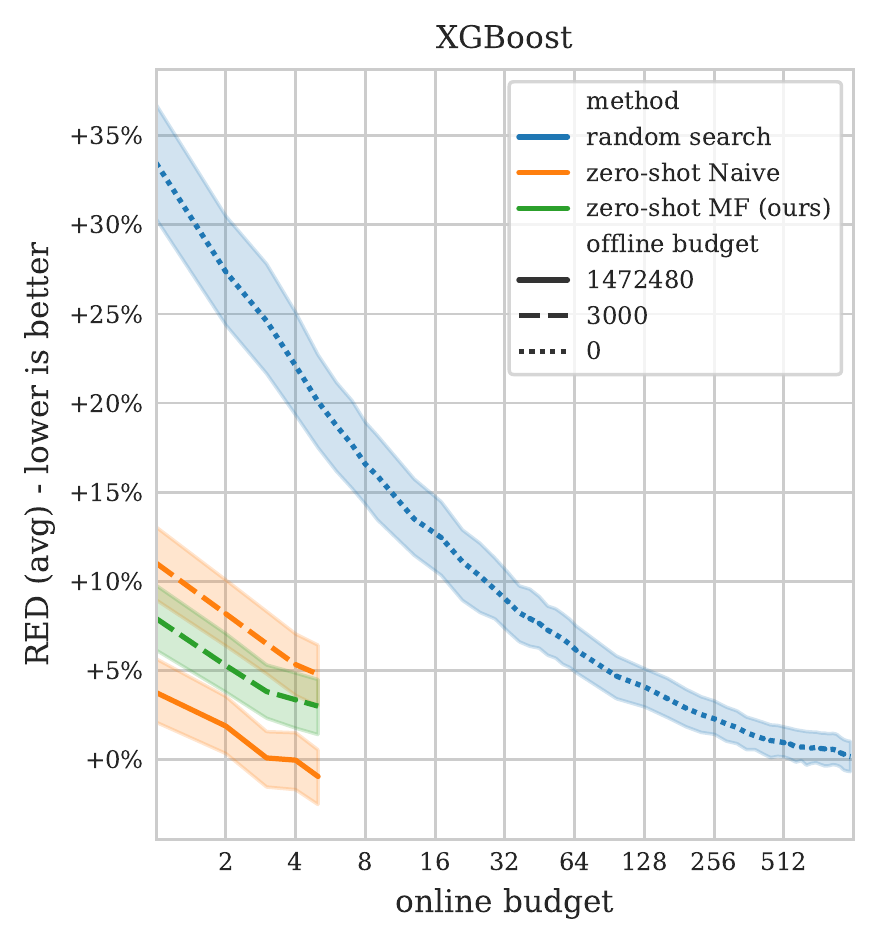}
  \end{subfigure}
  \begin{subfigure}[b]{0.40\linewidth}
    \includegraphics[width=\linewidth]{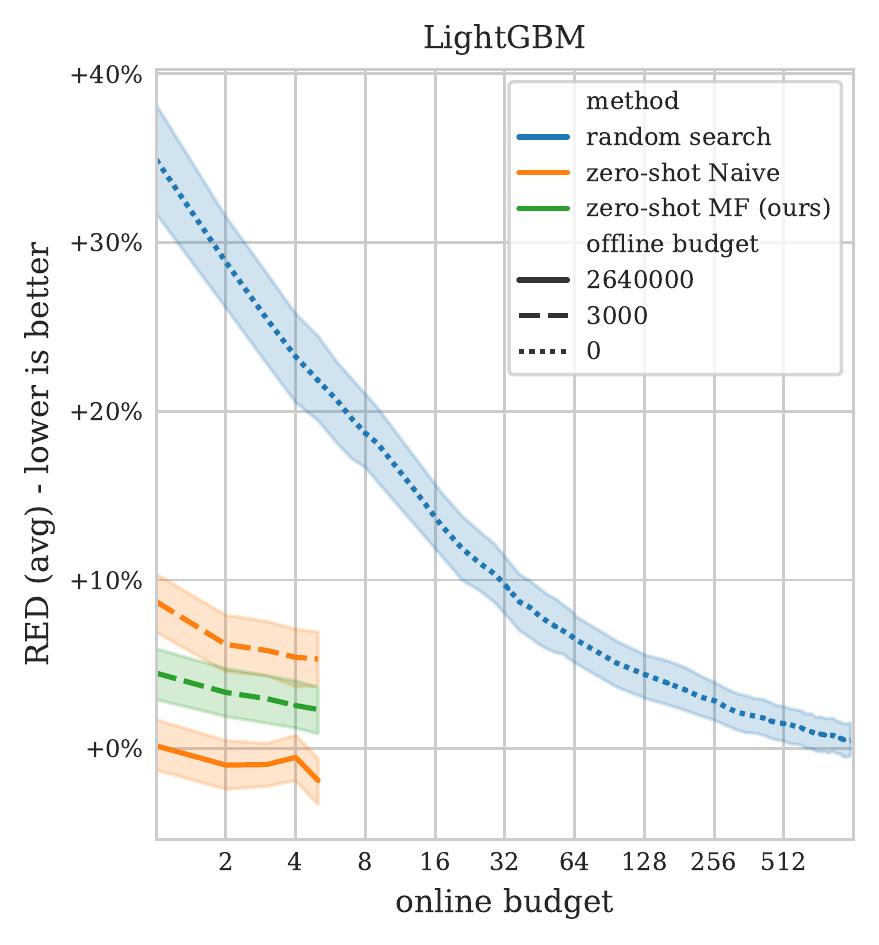}
  \end{subfigure}
  \begin{subfigure}[b]{0.40\linewidth}
    \includegraphics[width=\linewidth]{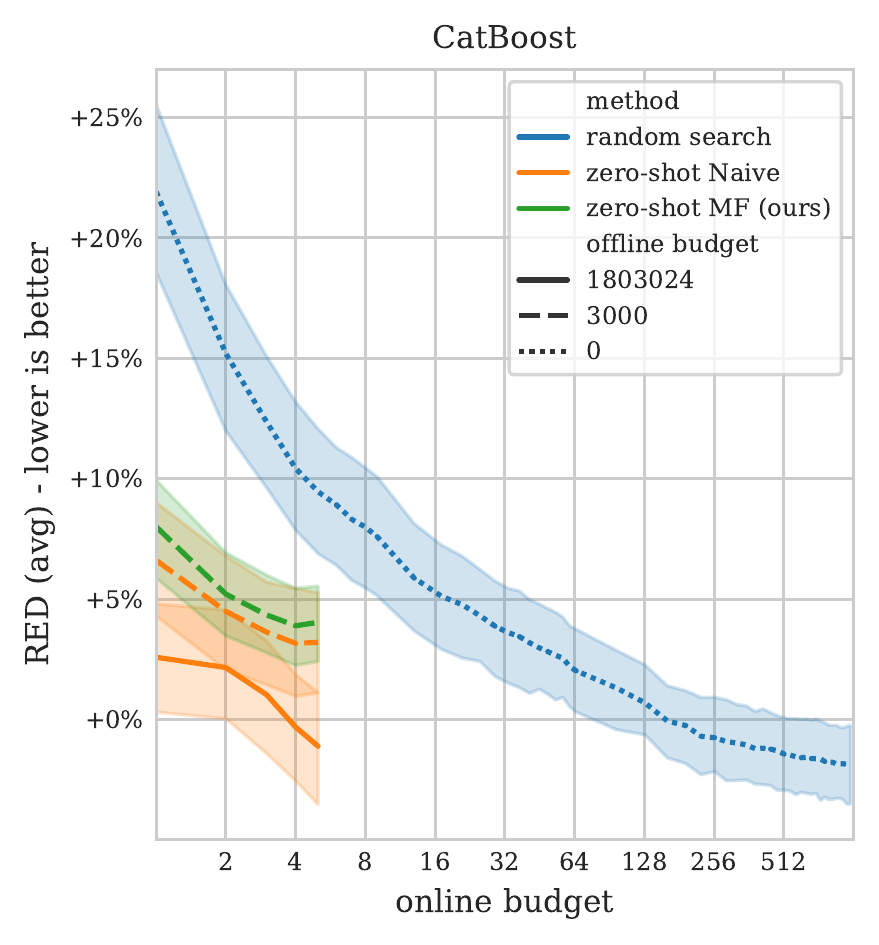}
  \end{subfigure}
  \begin{subfigure}[b]{0.40\linewidth}
    \includegraphics[width=\linewidth]{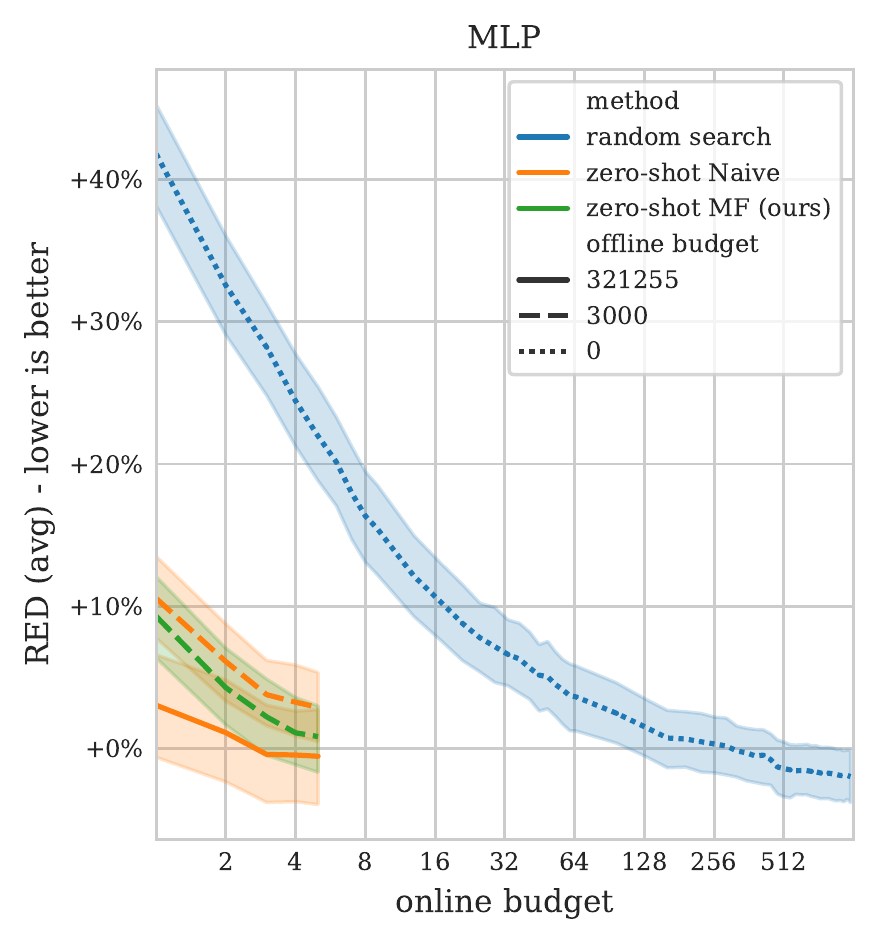}
  \end{subfigure}
  \begin{subfigure}[b]{0.40\linewidth}
    \includegraphics[width=\linewidth]{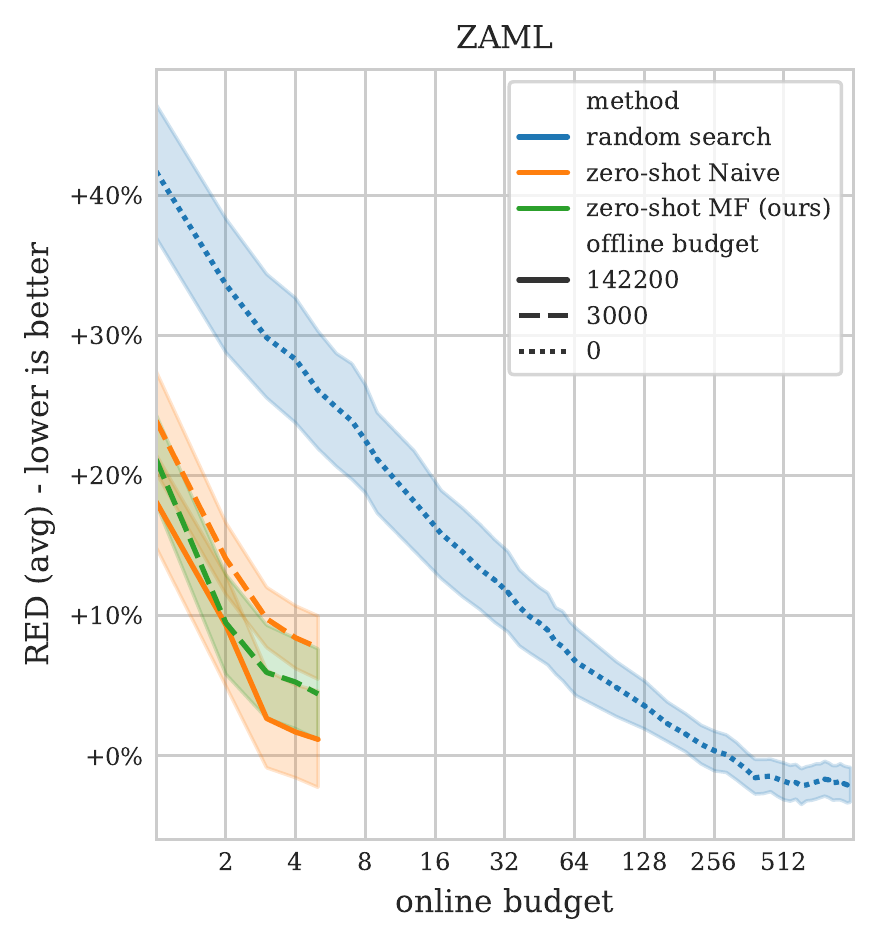}
  \end{subfigure}
  \caption{\textbf{Results for MF}. Performance obtained by performing HPO using zero-shot configurations found by our multi-fidelity algorithm (green). We compare the performance with the Naive baseline (orange) applied to all the data we have available (note the orders of magnitude more training jobs used to find these configurations) and on a randomly sub-sampled table using the same offline budget as the multi-fidelity algorithm. For comparison we also include the performance of random search, which does not require any offline training, but as expected requires a much larger online budget to find competitive HP configurations.}
  \label{fig:mf}
\end{figure}

We also provide additional plots of the results reported in the main text (Figure \ref{fig:obo} and \ref{fig:mf}). In all plots the RED for any given dataset is computed relative to the average test miss-classification rate of the 10 best configurations among all data we generated, selected by validation miss-classification rate. We plot the aggregate across all datasets, computed in a leave-one-dataset-out fashion. The shaded areas span one standard error in each direction, thus representing 68\% confidence intervals of the mean. We refer to the caption of the images for additional details.

\begin{figure}[ht]
\makebox[\linewidth][c]{
  \centering
  \begin{subfigure}[b]{0.218\linewidth}
    \includegraphics[width=\linewidth]{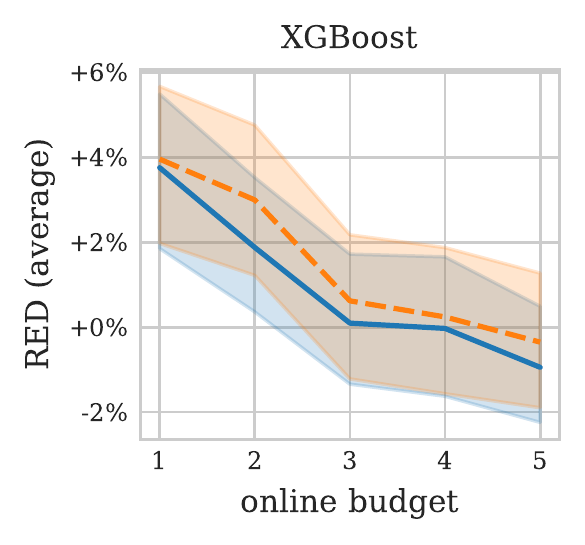}
  \end{subfigure}
  \begin{subfigure}[b]{0.20\linewidth}
    \includegraphics[width=\linewidth]{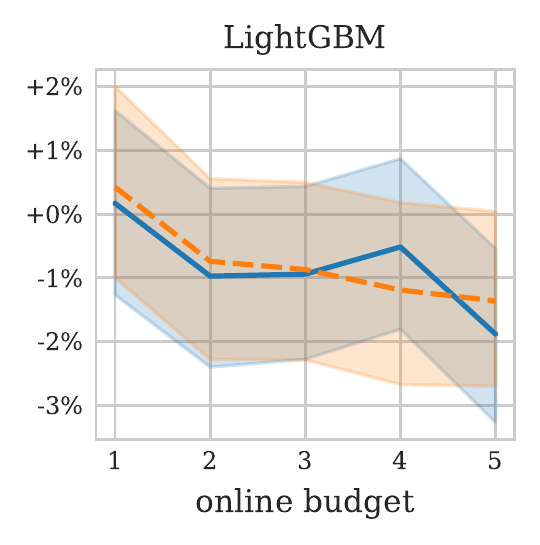}
  \end{subfigure}
  \begin{subfigure}[b]{0.20\linewidth}
    \includegraphics[width=\linewidth]{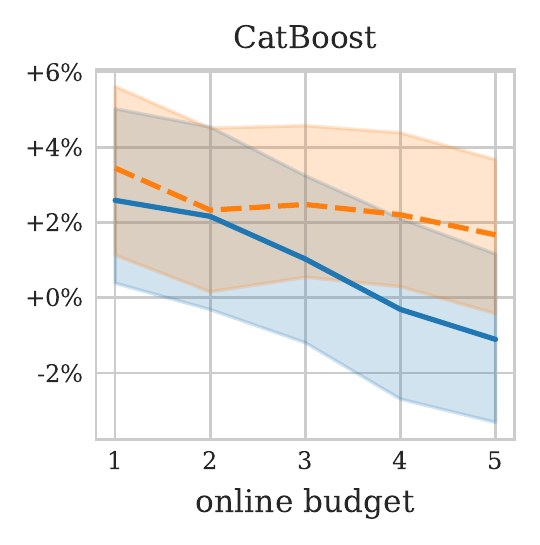}
  \end{subfigure}
  \begin{subfigure}[b]{0.20\linewidth}
    \includegraphics[width=\linewidth]{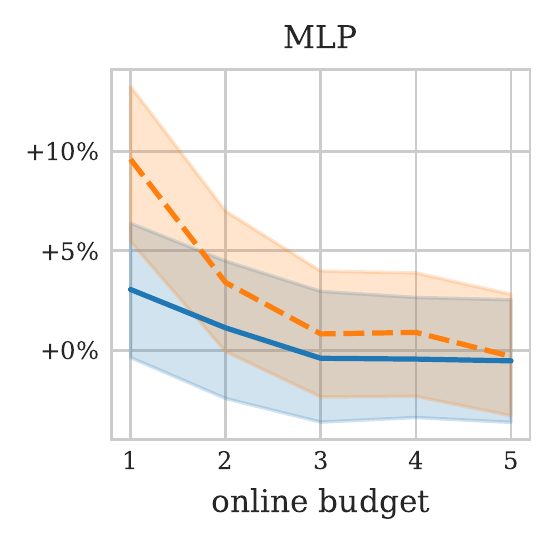}
  \end{subfigure}
  \begin{subfigure}[b]{0.20\linewidth}
    \includegraphics[width=\linewidth]{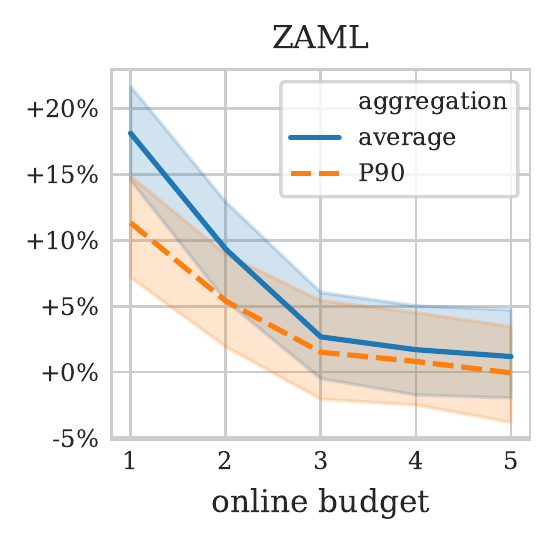}
  \end{subfigure}
 }
 
\makebox[\linewidth][c]{
  \centering
  \begin{subfigure}[b]{0.218\linewidth}
    \includegraphics[width=\linewidth]{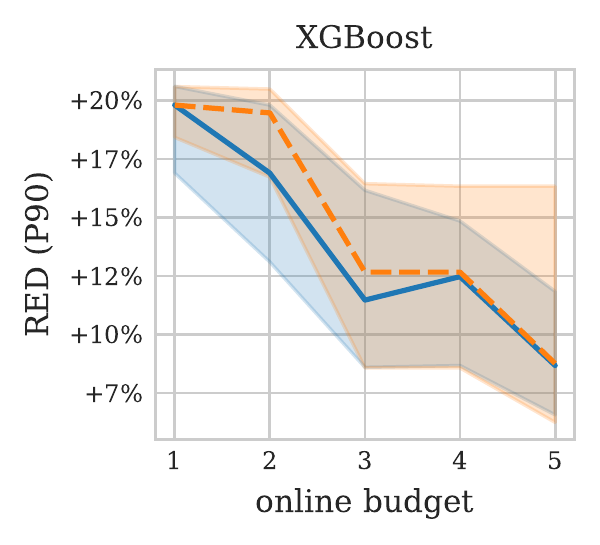}
  \end{subfigure}
  \begin{subfigure}[b]{0.20\linewidth}
    \includegraphics[width=\linewidth]{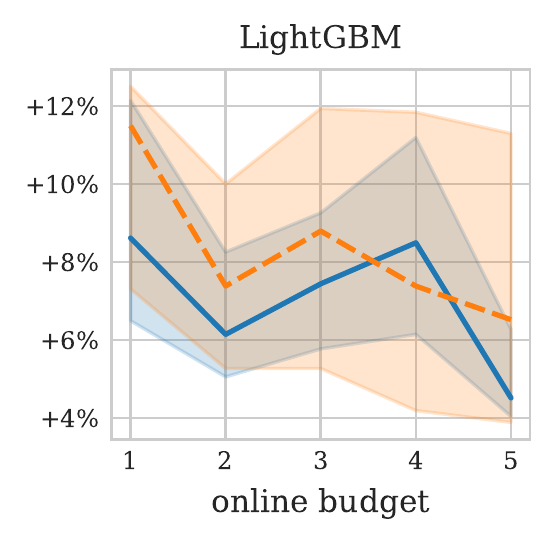}
  \end{subfigure}
  \begin{subfigure}[b]{0.20\linewidth}
    \includegraphics[width=\linewidth]{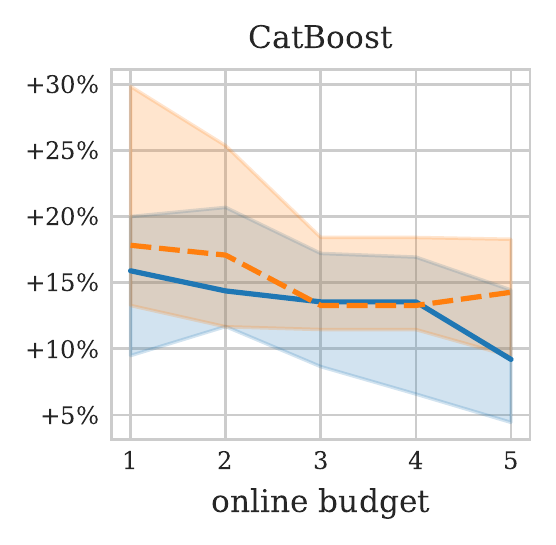}
  \end{subfigure}
  \begin{subfigure}[b]{0.20\linewidth}
    \includegraphics[width=\linewidth]{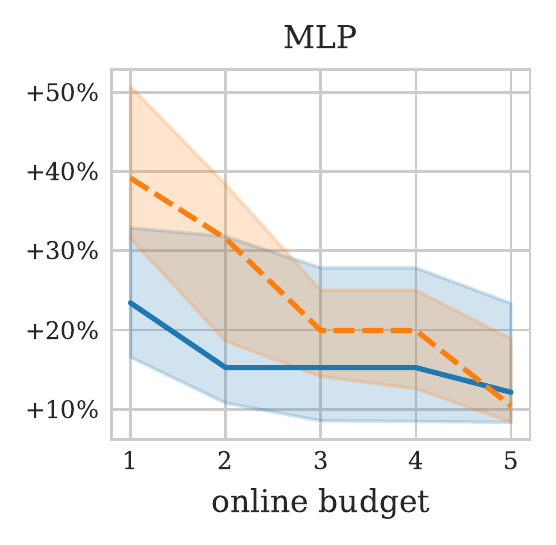}
  \end{subfigure}
  \begin{subfigure}[b]{0.20\linewidth}
    \includegraphics[width=\linewidth]{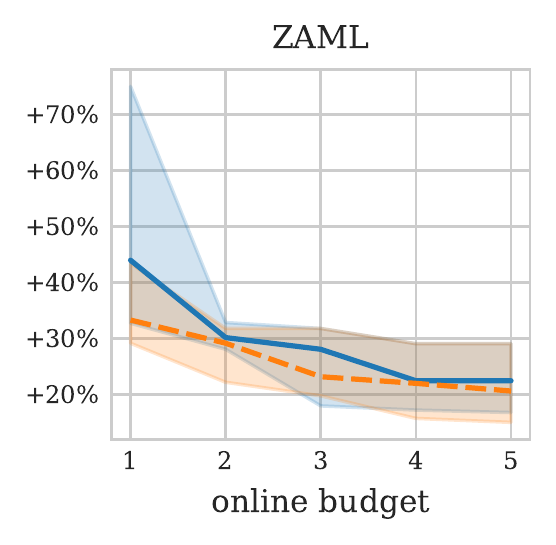}
  \end{subfigure}
  }
  \caption{\textbf{Ablation study: aggregation}. These plots compare the use of the \textbf{average} (blue) with the use of the \textbf{P90} (orange) when aggregating the meta-loss among different datasets. The top row shows the average performance of the zero-shot configurations found using the two strategies, while the bottom row shows the P90 across the datasets, with bootstrapped confidence intervals. }
  \label{fig:aggregation}
\end{figure}

Figure \ref{fig:aggregation} reports the result of an ablation study in which we compared the use of the average to that of the P90, when aggregating the meta-loss across all datasets. One reason one might be inclined to prefer the P90 is that it can be more robust to outliers and thus perform better on dataset dissimilar to those in our selection of datasets. However, our experiments show that using a P90 is unlikely to provide a benefit. Even if we are interested in the P90 (displayed in the second row of the figure) the zero-shot configurations computed using the average match -- and in most cases outperform -- the configurations computed by optimizing the P90.

\begin{figure}[ht]
\makebox[\linewidth][c]{
  \centering
  \begin{subfigure}[b]{0.218\linewidth}
    \includegraphics[width=\linewidth]{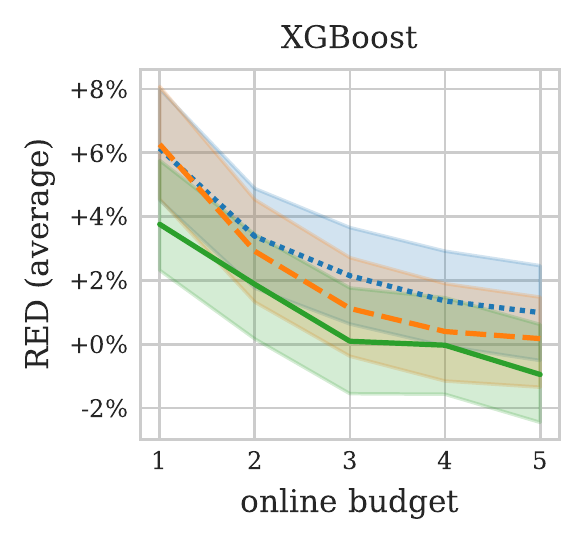}
  \end{subfigure}
  \begin{subfigure}[b]{0.20\linewidth}
    \includegraphics[width=\linewidth]{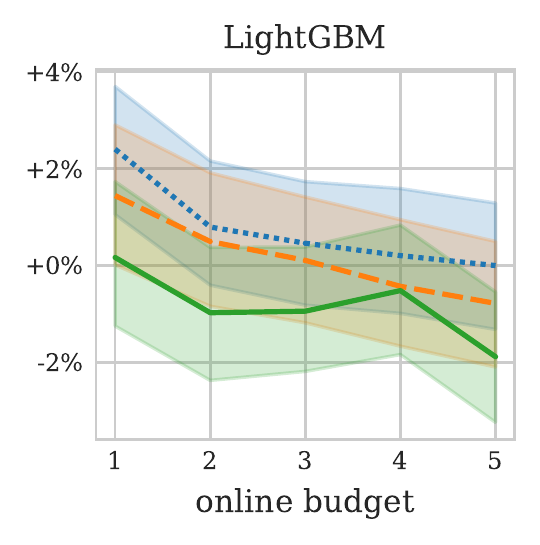}
  \end{subfigure}
  \begin{subfigure}[b]{0.20\linewidth}
    \includegraphics[width=\linewidth]{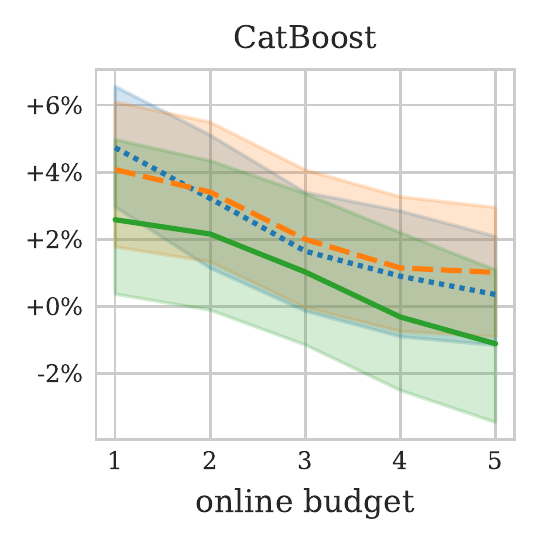}
  \end{subfigure}
  \begin{subfigure}[b]{0.20\linewidth}
    \includegraphics[width=\linewidth]{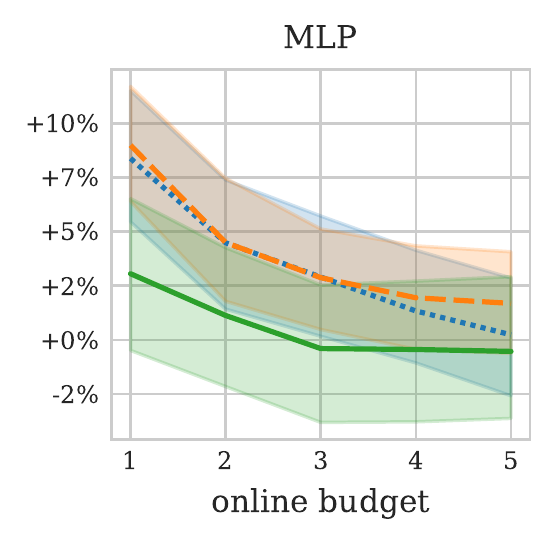}
  \end{subfigure}
  \begin{subfigure}[b]{0.20\linewidth}
    \includegraphics[width=\linewidth]{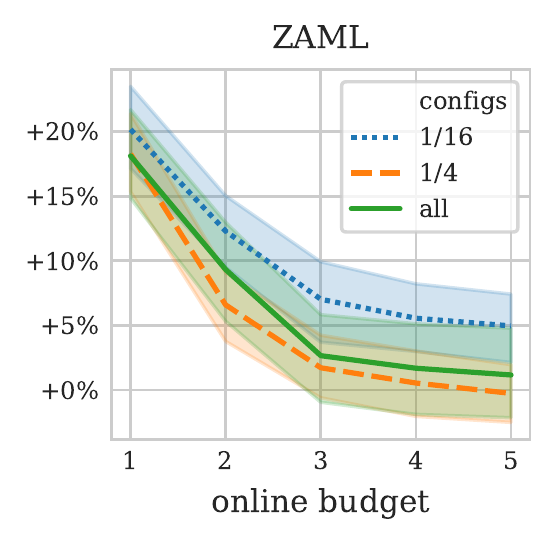}
  \end{subfigure}
  }
  \caption{\textbf{Ablation study: number of configurations}. These plots show the effect of reducing the number of configurations considered. In all cases we use the Naive algorithm. We can see that in general using the full data we generated (green) provides better zero-shot configurations than using 1/4th of the data (orange) and 1/16th of the data (blue).}
  \label{fig:configs}
\end{figure}

Finally, Figure \ref{fig:configs} shows the effect of the number of random configurations considered. These experiments show that considering more configurations does in fact allow us to compute better zero-shot configurations, thus providing a good argument in favour of more sample efficient zero-shot HPO algorithms.

\newpage

\begin{table}[tbhp]
\centering
\caption{OBO (vs MF) -- avg RED ($\pm$std. err.) -- lower is better.}\label{tab:obo mf}
\begin{tabular}{lrrr}
\toprule
 & 1 zero-shot config                   & 2 zero-shot configs           & 5 zero-shot configs     \\
\midrule
\textbf{XGBoost} & -2.44\% ($\pm$ 0.87) & -2.78\% ($\pm$ 0.79) & -2.75\% ($\pm$ 0.64) \\
\textbf{LightGBM} & -4.37\% ($\pm$ 1.41) & -4.72\% ($\pm$ 1.38) & -4.68\% ($\pm$ 1.32) \\
\textbf{CatBoost} & -3.32\% ($\pm$ 0.98) & -1.47\% ($\pm$ 0.79) & -2.24\% ($\pm$ 0.90) \\
\textbf{MLP} & -1.52\% ($\pm$ 1.50) & -2.01\% ($\pm$ 1.20) & -1.46\% ($\pm$ 1.40) \\
\textbf{ZAML} & -0.15\% ($\pm$ 2.64) & +1.68\% ($\pm$ 2.49) & +0.18\% ($\pm$ 2.66) \\
\bottomrule 
\textbf{Combined} & -2.70\% ($\pm$ 0.63)  & -2.35\% ($\pm$ 0.58)  & -2.57\% ($\pm$ 0.59)  \\
\end{tabular}

\centering
\caption{Surrogate (vs Naive) -- avg RED ($\pm$std. err.) -- lower is better.}\label{tab:surr naive}
\begin{tabular}{lrrr}
\toprule
 & 1 zero-shot config                   & 2 zero-shot configs           & 5 zero-shot configs     \\
\midrule
\textbf{XGBoost} & -1.98\% ($\pm$ 1.87) & -3.09\% ($\pm$ 1.81) & -4.15\% ($\pm$ 1.75) \\
\textbf{LightGBM} & -7.42\% ($\pm$ 1.76) & -6.46\% ($\pm$ 1.67) & -6.67\% ($\pm$ 1.62) \\
\textbf{CatBoost} & +0.04\% ($\pm$ 1.82) & +0.61\% ($\pm$ 1.72) & +0.99\% ($\pm$ 1.65) \\
\textbf{MLP} & +6.04\% ($\pm$ 3.05) & -0.11\% ($\pm$ 1.36) & -2.68\% ($\pm$ 1.33) \\
\textbf{ZAML} & -0.36\% ($\pm$ 2.34) & +1.53\% ($\pm$ 1.89) & -0.81\% ($\pm$ 1.75) \\
\bottomrule 
\textbf{Combined} & -1.38\% ($\pm$ 0.98)  & -2.11\% ($\pm$ 0.80)  & -3.06\% ($\pm$ 0.77)  \\
\end{tabular}

\centering
\caption{MF (vs Surrogate) -- avg RED ($\pm$std. err.) -- lower is better.}\label{tab:mf surr}
\begin{tabular}{lrrr}
\toprule
 & 1 zero-shot config                   & 2 zero-shot configs           & 5 zero-shot configs     \\
\midrule
\textbf{XGBoost} & -2.90\% ($\pm$ 0.81) & -2.11\% ($\pm$ 0.78) & -0.16\% ($\pm$ 0.67) \\
\textbf{LightGBM} & +3.66\% ($\pm$ 1.35) & +3.16\% ($\pm$ 1.27) & +3.54\% ($\pm$ 1.25) \\
\textbf{CatBoost} & +2.22\% ($\pm$ 1.27) & +1.35\% ($\pm$ 1.04) & +1.08\% ($\pm$ 1.07) \\
\textbf{MLP} & -5.70\% ($\pm$ 3.24) & -1.17\% ($\pm$ 1.77) & +0.59\% ($\pm$ 1.56) \\
\textbf{ZAML} & +1.38\% ($\pm$ 3.58) & -0.44\% ($\pm$ 3.34) & +1.75\% ($\pm$ 3.02) \\
\bottomrule 
\textbf{Combined} & -0.09\% ($\pm$ 0.87)  & +0.35\% ($\pm$ 0.67)  & +1.42\% ($\pm$ 0.62)  \\
\end{tabular}

\centering
\caption{OBO (vs Naive) -- avg RED ($\pm$std. err.) -- lower is better.}\label{tab:obo naive}
\begin{tabular}{lrrr}
\toprule
 & 1 zero-shot config                   & 2 zero-shot configs           & 5 zero-shot configs     \\
\midrule
\textbf{XGBoost} & -7.21\% ($\pm$ 1.90) & -7.96\% ($\pm$ 1.85) & -7.14\% ($\pm$ 1.82) \\
\textbf{LightGBM} & -7.69\% ($\pm$ 1.75) & -6.92\% ($\pm$ 1.68) & -7.33\% ($\pm$ 1.64) \\
\textbf{CatBoost} & -1.87\% ($\pm$ 1.17) & -0.60\% ($\pm$ 1.15) & -1.40\% ($\pm$ 0.84) \\
\textbf{MLP} & -2.85\% ($\pm$ 2.01) & -3.47\% ($\pm$ 1.57) & -2.78\% ($\pm$ 1.84) \\
\textbf{ZAML} & -2.77\% ($\pm$ 3.48) & -2.90\% ($\pm$ 3.32) & -3.52\% ($\pm$ 3.04) \\
\bottomrule 
\textbf{Combined} & -4.92\% ($\pm$ 0.88)  & -4.75\% ($\pm$ 0.83)  & -4.78\% ($\pm$ 0.80)  \\
\end{tabular}
\end{table}

\end{document}